\documentclass{article} % For LaTeX2e
\usepackage{iclr2027_conference,times}
\makeatletter
\renewcommand{\@evenhead}{}
\renewcommand{\@oddhead}{}
\makeatother
\usepackage{amsmath,amsfonts,bm}

\def\eqref#1{equation~\ref{#1}}
\def\1{\bm{1}}

\DeclareMathAlphabet{\mathsfit}{\encodingdefault}{\sfdefault}{m}{sl}
\SetMathAlphabet{\mathsfit}{bold}{\encodingdefault}{\sfdefault}{bx}{n}

\usepackage{hyperref}
\usepackage{url}
\usepackage{graphicx}

\title{Where to Compute and How to Interact: Operator-Readable Adaptation 
with Gauge-Aware Transport}
\iclrfinalcopy
\author{
Zixuan Shen$^{1*}$ \quad
Quanxu Wan$^{1*}$ \quad
Bingchuan Wang$^{1}$ \quad
Zhi Wang$^{2}$ \quad
Biao Luo$^{1}$\\
$^{1}$Central South University \quad
$^{2}$Nanjing University
}

\begin{document}

\maketitle

\begin{abstract}
Adaptive meshes enable neural operators for partial differential equations (PDEs) to allocate spatial samples and computational resources according to local physical structures. Existing approaches, however, primarily focus on where to compute, while paying less attention to how to interact after node relocation. Mesh adaptation changes local sampling scales, neighborhood structures, and geometric contexts, making representations formed at different nodes not necessarily directly comparable. Direct aggregation may therefore entangle genuine physical variation with discretization-induced representation variation. Moreover, because allocation and interaction are jointly optimized through the same output objective, their individual roles are difficult to distinguish from final errors alone. We introduce operator readability, which requires an adaptive operator to explicitly account for and test why computation is allocated to particular locations and how representations interact under the resulting nonuniform discretization. Based on this principle, we propose the Gauge-Aware Adaptive Mesh Neural Operator (GA-AMNO). Physics-informed adaptive allocation answers where to compute, while geometry-conditioned low-rank Gauge transport maps source features into target representation contexts before aggregation, answering how to interact. This design turns the otherwise implicit mesh-to-solver information exchange into an inspectable and intervenable computational process. We further establish sufficient conditions for representation-consistent aggregation and analyze approximate transport errors and continuity under topology-preserving mesh deformations. Experiments across five PDE benchmarks demonstrate improved predictive accuracy, while controlled interventions and geometric-mismatch analyses verify the computational roles of allocation and interaction and show that Gauge transport improves cross-discretization representation compatibility under strong geometric mismatch.
\end{abstract}

\section{Introduction}

Neural operators aim to learn solution mappings for partial differential equations (PDEs) between function spaces, while their numerical realizations operate on physical fields sampled over grids or meshes \citep{huang2025pde,kovachki2023neuraloperator}. Adaptive discretization makes this computational representation state dependent by relocating nodes toward dynamically informative regions \citep{berger1984adaptive}. Existing adaptive operators are therefore commonly organized around one question: where should computation be allocated? However, node relocation changes not only the spatial distribution of resolution, but also the local sampling scales, edge directions, neighborhood structures, and mesh Jacobians under which features are formed. The resulting representations must subsequently be exchanged by the operator under these unequal local geometric conditions. Adaptive computation must therefore answer a second question: how should these representations interact? These two decisions are connected through the learned geometry. Allocation determines the local discretization contexts presented to the interaction mechanism, while interaction determines whether the representations created by that allocation can be combined effectively. Making the mesh adaptive without explicitly accounting for the resulting information exchange consequently leaves an important part of adaptive computation unspecified.

\begin{figure*}[t]
\centering
\includegraphics[width=0.98\textwidth]
{\detokenize{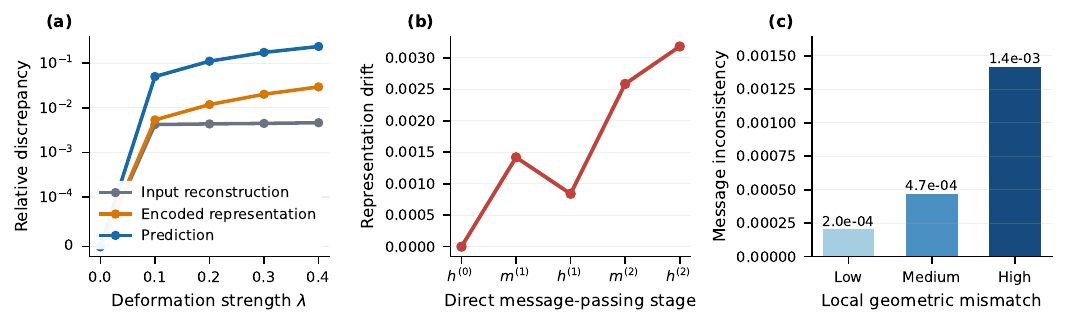}}
\caption{Preliminary diagnosis of the interaction effects induced by
adaptive discretization on Navier--Stokes.
(a) For the same physical state, increasing the topology-preserving
deformation strength $\lambda$ produces substantially larger representation
and prediction discrepancies than input reconstruction discrepancy.
(b) Representation drift emerges at direct aggregation and accumulates
through subsequent message passing, where $\mathbf{h}^{(\ell)}$ denotes the
node representation at layer $\ell$ and $\mathbf{m}^{(\ell)}$ the aggregated
message.
(c) Message inconsistency increases with local geometric mismatch.
Together, these results show that direct aggregation amplifies differences
induced by unequal local discretization contexts.}
\label{fig:preliminary_diagnosis}
\end{figure*}

This missing account of information exchange is difficult to expose under
standard end-to-end training, because allocation and interaction are jointly
optimized through the same prediction objective. An expressive solver may
compensate for an uninformative mesh, while a strong allocation mechanism may
reduce the apparent need for geometry-aware interaction. Different
allocation--interaction configurations may therefore achieve similar
predictive performance, even though they use the adaptive geometry in
substantially different ways. To determine whether this concern has observable
consequences, we conduct a controlled diagnosis in
\autoref{fig:preliminary_diagnosis}. We keep the physical state, model
parameters, node identities, and neighborhood topology fixed, while
progressively deforming only the sampling geometry. A small input
reconstruction discrepancy develops into substantially larger representation
and prediction discrepancies. Layerwise measurements further locate the onset
of this drift at direct message aggregation, while message inconsistency
increases with local geometric mismatch. These results show that the problem
is not irregular sampling alone; it emerges when representations formed under
unequal local discretization contexts are directly combined.

These observations motivate operator readability as an operational and experimentally testable computational criterion. An operator-readable adaptive model should make the roles of its adaptive decisions explicit and allow each role to be examined through controlled interventions. In this work, we use the following conditions as sufficient operational criteria for the two aspects of readability. \emph{Allocation readability} is established when node movement is explicitly linked to declared physical quantities through a clear computational path and responds predictably to controlled modifications of the allocation signal. \emph{Interaction readability} is established when source-to-target information transformation is explicitly conditioned on local geometric relations, can be examined at the edge level, and exhibits measurable responses to geometric mismatch and controlled interventions. Readability therefore does not refer to whether the resulting mesh is visually intuitive. It refers to whether the model can account for both why computational resolution is allocated to particular regions and how information is exchanged under the resulting discretization.

\begin{figure*}[t]
\centering
\includegraphics[width=0.98\textwidth]
{\detokenize{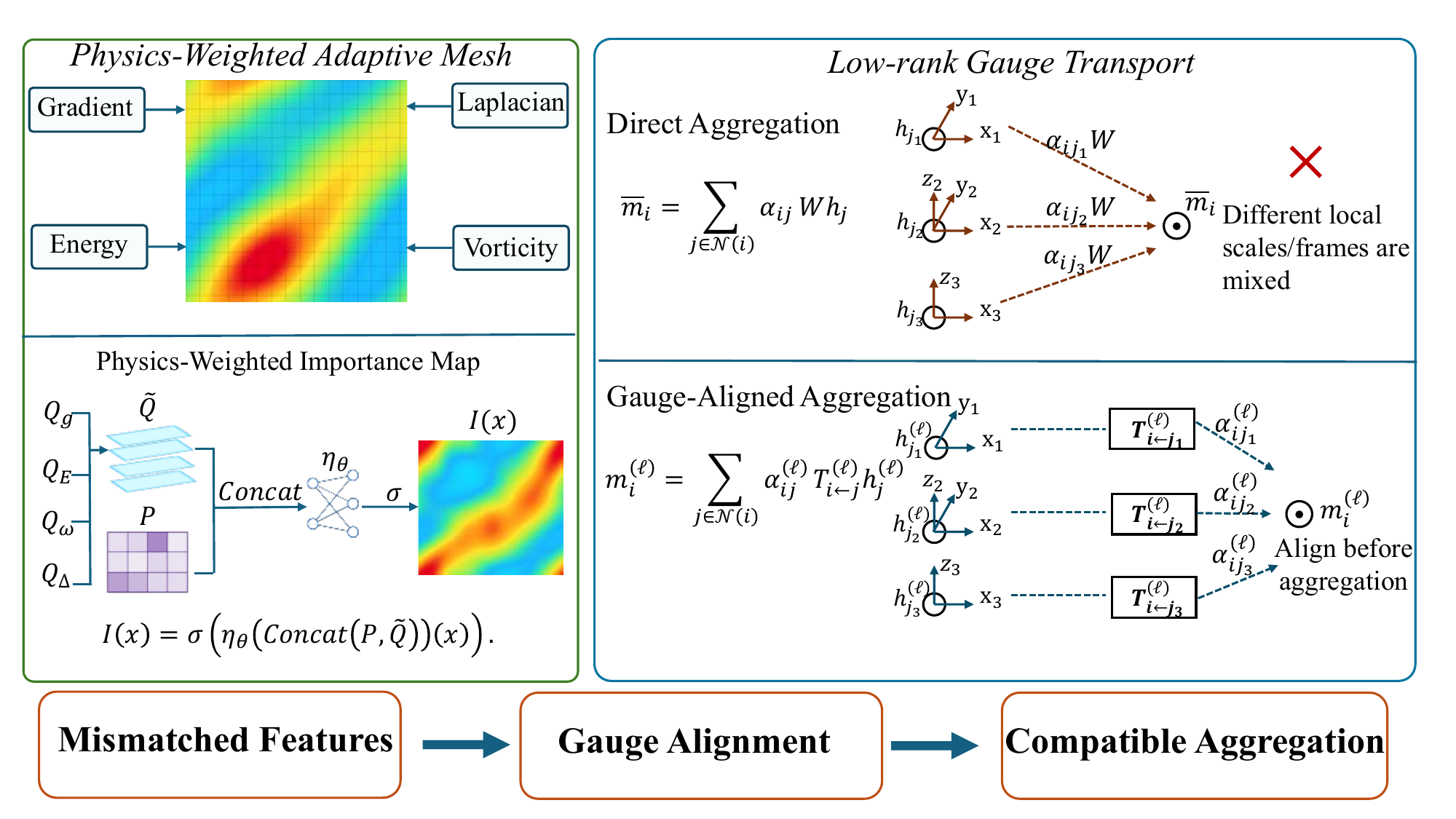}}
\caption{Conceptual overview of operator-readable adaptive computation.
Physics-informed allocation explains where adaptive resolution is placed.
The resulting nonuniform mesh produces unequal local discretization contexts,
under which direct aggregation leaves the effects of geometric mismatch on
cross-node interaction implicit. Gauge-aware transport instead maps each
source feature into a target-conditioned representation before aggregation,
making explicit how information is exchanged after the discretization changes.}
\label{fig:gauge_concept}
\end{figure*}

Based on this formulation, we propose the Gauge-Aware Adaptive Mesh
Neural Operator (GA-AMNO) as one realization of operator-readable
adaptation. As summarized in \autoref{fig:gauge_concept}, physics-informed
indicators construct an importance map that determines where nodes are
relocated, while the resulting relative coordinates, local mesh
Jacobians, and state descriptors condition an edge-wise low-rank
transport that determines how their features interact. We use Gauge
transport as a design principle rather than assume that every
discretization change induces an exact gauge transformation. By mapping
each source feature into its target representation context before
aggregation, it explicitly accounts for differences in local sampling
scale, orientation, and neighborhood geometry, making the otherwise
implicit mesh-to-solver interaction inspectable. Gauge transport
therefore provides a structured realization of interaction readability,
while the predictive capability is supplied by the complete GA-AMNO
architecture. We further establish sufficient conditions for
representation-consistent aggregation, characterize the propagation of
approximate transport error, and analyze continuity under
topology-preserving mesh deformation.

The main contributions are summarized as follows:
\begin{itemize}
    \item \textbf{Conceptual contribution.}
    We reformulate adaptive neural operator learning as two connected problems: resource allocation and cross-discretization interaction. Based on this formulation, we introduce an operational notion of operator readability and identify, through controlled re-discretization, a reproducible failure mode in which direct aggregation amplifies discretization-induced representation drift.

    \item \textbf{Architectural and theoretical contribution.}
    We develop GA-AMNO as a concrete realization of operator-readable adaptation. Physics-informed allocation determines where computational resolution is placed, while geometry-conditioned low-rank transport makes explicit how representations formed under unequal local discretizations interact. We additionally provide consistency conditions, an approximate-transport error bound, and a continuity analysis under topology-preserving mesh deformation.

    \item \textbf{Evaluation contribution.}
    We evaluate both predictive utility and mechanism readability across multiple PDE families. Beyond standard prediction benchmarks, allocation counterfactuals, edge interventions, geometric-mismatch stratification, and controlled mesh deformations test whether the allocation and interaction mechanisms fulfill their declared computational roles.
\end{itemize}

\section{Related Work}
\label{sec:related_work}

We briefly review the three lines of research most relevant to our work. A more comprehensive discussion is provided in \autoref{sec:expanded_related_work}.

\textbf{Neural operators.}
Neural operators learn mappings between function spaces using spectral transformations, coordinate-conditioned kernels, graph interactions, or physics-aware attention
\citep{li2020fourier,li2020neural,wu2024transolver,bie2025space}.
While these methods support regular or irregular discretizations, they do not primarily examine how input-dependent mesh adaptation changes the representation contexts in which features interact.

\textbf{Adaptive mesh methods.}
Classical and learned adaptive methods allocate spatial resolution through error indicators, local refinement, node movement, or learned policies
\citep{berger1984adaptive,berger1989local,rudd2014adaptive,pfaff2020learning}.
They mainly determine where computation should be concentrated, whereas the interaction between features formed under the resulting nonuniform local discretizations is typically delegated to the downstream solver.

\textbf{Gauge-aware representations.}
Gauge-aware geometric networks transform features between local reference frames to support consistent information exchange
\citep{bronstein2021geometric,cohen2019gauge}.
Motivated by this principle, we use geometry-conditioned transport without assuming exact Gauge equivariance, making feature interaction across adaptive local discretizations explicit before aggregation.

\section{Method}

\subsection{Problem Formulation}
\label{sec:problem_formulation}

Let $\Omega \subset \mathbb{R}^{d}$ be the spatial domain and let
$\mathbf{u}$ denote a $C$-channel physical field. We consider PDE systems that
can be written in the generic form
\begin{equation}
    \mathcal{F}_{\mathrm{PDE}}
    \left(
        \mathbf{u},
        \partial_t\mathbf{u},
        \nabla\mathbf{u},
        \nabla^2\mathbf{u};
        \boldsymbol{\mu}
    \right)
    =0.
\end{equation}
Here, $\mathcal{F}_{\mathrm{PDE}}$ is the governing differential operator, $\partial_t$ is the temporal derivative, $\nabla$ and $\nabla^2$ denote first- and second-order spatial derivatives, and $\boldsymbol{\mu}$ collects physical parameters, coefficient fields, or forcing terms. For steady-state PDEs, the temporal derivative is omitted.

The corresponding solution operator is
\begin{equation}
    \mathcal{G}^{\dagger}:\mathcal{A}\rightarrow\mathcal{U},
    \qquad
    \mathbf{u}=\mathcal{G}^{\dagger}(\mathbf{a}),
\end{equation}
where $\mathcal{A}$ is the input function space, $\mathcal{U}$ is the solution function space, $\mathbf{a}$ denotes the input condition, and $\mathcal{G}^{\dagger}$ is the unknown exact operator \cite{kovachki2023neuraloperator}. For temporal prediction, $\mathbf{a}$ contains a short history of states; for steady-state prediction, it may contain coefficients, forcing terms, or boundary information.

GA-AMNO learns an approximation
\begin{equation}
    \widehat{\mathbf{u}}
    =
    \mathcal{G}_{\theta}(\mathbf{a})
    \approx
    \mathcal{G}^{\dagger}(\mathbf{a}),
\end{equation}
where $\widehat{\mathbf{u}}$ is the predicted field and $\mathcal{G}_{\theta}$ is the neural operator with trainable parameters $\theta$.

The input field is first represented on a reference discretization
$\mathcal{X}^{r}=\{\mathbf{x}^{r}_{i}\}_{i=1}^{N}$. GA-AMNO then constructs an input-dependent adaptive mesh
\begin{equation}
    \mathcal{X}^{a}
    =
    \left\{\mathbf{x}^{a}_{i}=\Pi_{\Omega}\left(\mathbf{x}^{r}_{i}+\Delta\mathbf{x}_{i}\right)\right\}_{i=1}^{N}.
\end{equation}
Here, $\mathcal{X}^{a}$ is the adaptive mesh, $\mathbf{x}^{r}_{i}$ and $\mathbf{x}^{a}_{i}$ are the reference and adaptive coordinates of node $i$, $N$ is the number of nodes, $\Delta\mathbf{x}_{i}\in\mathbb{R}^{d}$ is the learned displacement, and $\Pi_{\Omega}$ keeps the displaced node inside the computational domain.

The adaptive coordinates define an input-dependent local discretization
environment for each node. Once constructed for an input $\mathbf{a}$, the
adaptive mesh $\mathcal{X}^{a}(\mathbf{a})$ remains fixed during the subsequent
operator computation but may vary across inputs, leading to changes in local
spacing, neighborhood geometry, and mesh Jacobians. Consequently, nodal
features describing related physical content may be formed under different
local discretization environments and need not be directly comparable. To address this mismatch, GA-AMNO does not assume an analytically specified coordinate transformation.
Instead, it learns local feature-transport mappings that are conditioned on
mesh geometry and input descriptors and are applied before aggregation to mitigate
the resulting representation offsets.

\subsection{Overall Architecture}
\label{sec:overall_architecture}

\begin{figure*}[t]
\centering
\includegraphics[width=0.98\textwidth]{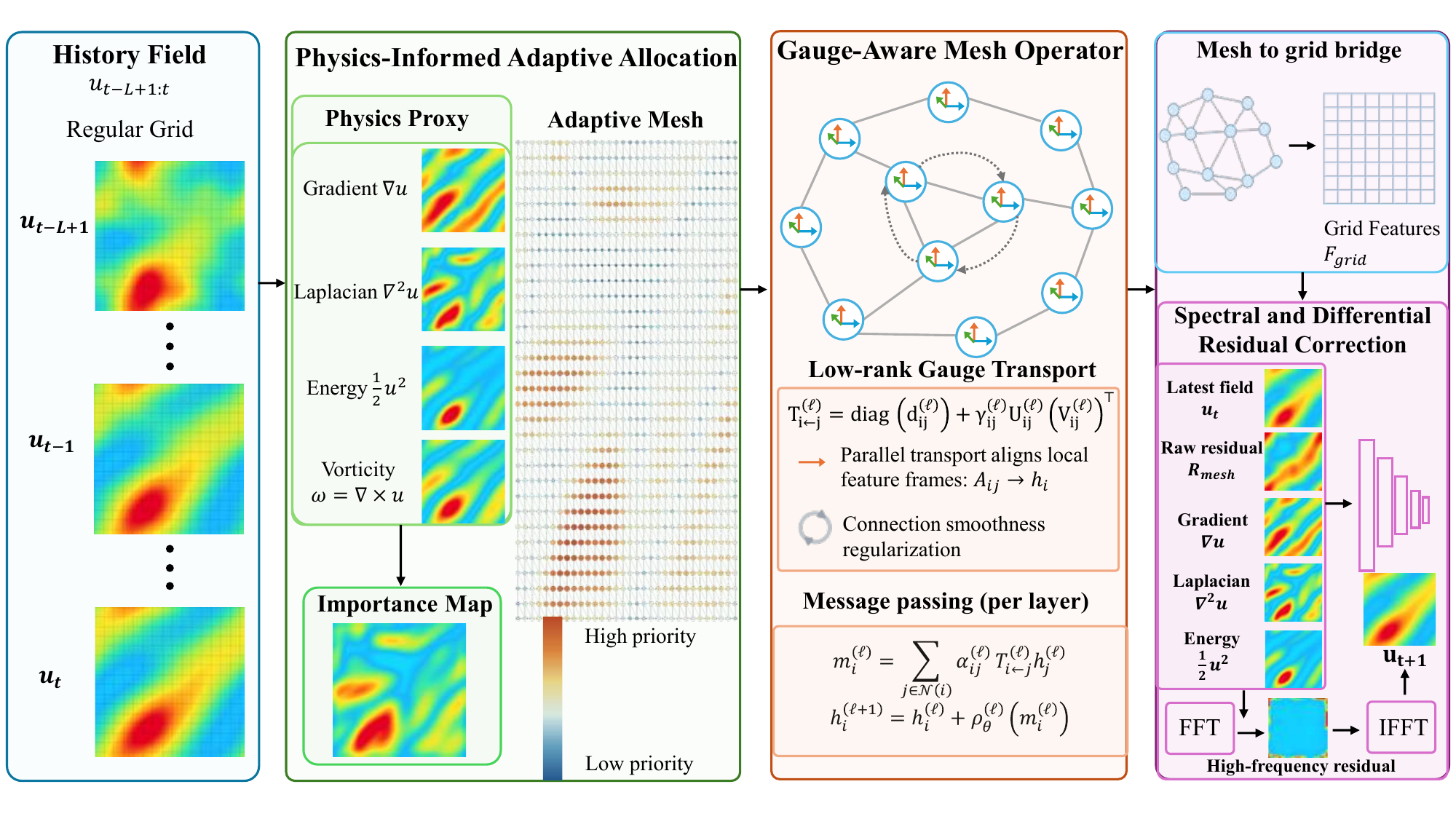}
\caption{Overall architecture of GA-AMNO. Physics-informed indicators guide adaptive allocation, low-rank gauge transport aligns features before aggregation on the resulting mesh, and the reconstructed grid features are refined by spectral and differential residual corrections.}
\label{fig:ga_amno_architecture}
\end{figure*}

Given an input $\mathbf{a}$ on $\mathcal{X}^{r}$, GA-AMNO encodes local state
descriptors and physics-informed indicators, predicts an importance map, and
constructs $\mathcal{X}^{a}(\mathbf{a})$. Gauge-aware operator layers then use
adaptive coordinates, local mesh Jacobians, and state descriptors to transport
each source feature into its target context before aggregation. After the nodal
updates, a mesh-to-grid bridge reconstructs the adaptive features on the
reference grid, where spectral and differential residual corrections produce
the final output, as summarized in \autoref{fig:ga_amno_architecture}. The complete forward procedure is given in
\autoref{app:algorithm}, while the reconstruction details are provided in
\autoref{sec:mesh_to_grid_bridge}.

\subsection{Physics-Informed Adaptive Allocation}
\label{sec:adaptive_mesh}

This module determines where representation capacity is allocated. Given the
latest input slice $\mathbf{v}$, we compute four physically meaningful
indicators: the gradient magnitude $Q_g$, Laplacian magnitude $Q_{\Delta}$,
local energy $Q_E$, and vorticity-related response $Q_{\omega}$. Their complete
definitions and scalar-field treatment are provided in
Appendix~\ref{app:adaptive_allocation_details}. The indicators are combined
with the learned spatial state descriptor
$\mathbf{P}=\operatorname{Encoder}(\mathbf{a})$ to predict the importance map:
\begin{align}
    \widetilde{\mathbf{Q}}(\mathbf{x})
    &=
    \operatorname{Concat}\!\left(
    \mathcal{N}(Q_g),
    \mathcal{N}(Q_{\Delta}),
    \mathcal{N}(Q_E),
    \mathcal{N}(Q_{\omega})
    \right)(\mathbf{x}),
    \nonumber\\
    I(\mathbf{x})
    &=
    \sigma\!\left(
    \eta_{\theta}\!\left(
    \operatorname{Concat}(\mathbf{P},\widetilde{\mathbf{Q}})
    \right)(\mathbf{x})
    \right).
    \label{eq:importance_prediction}
\end{align}
Here, $\mathbf{x}$ is a spatial location,
$\widetilde{\mathbf{Q}}(\mathbf{x})$ is the concatenated physical-indicator
vector at $\mathbf{x}$, $\mathcal{N}(\cdot)$ denotes per-sample channel
normalization, and $\operatorname{Concat}(\cdot)$ denotes channel-wise
concatenation. The quantity $I(\mathbf{x})$ is the scalar importance value,
$\eta_{\theta}$ is a convolutional predictor parameterized by $\theta$, and
$\sigma$ is the sigmoid function. The input $\mathbf{a}$ and its encoded
descriptor $\mathbf{P}$ have been defined above.

The predicted importance modulates a bounded node displacement:
\begin{align}
    \mathbf{o}_{i}
    &=
    \tanh\!\left(\psi_{\theta}(\mathbf{P})_i\right),
    \nonumber\\
    w_i
    &=
    \frac{I_{\min}+s_I I_i}
    {N^{-1}\sum_{k=1}^{N}(I_{\min}+s_I I_k)+\epsilon},
    \nonumber\\
    \Delta\mathbf{x}_{i}
    &=
    \delta_{\max}w_i\mathbf{o}_{i}.
    \label{eq:adaptive_displacement}
\end{align}
Here, $\mathbf{o}_i$ is the bounded raw displacement of node $i$,
$\psi_{\theta}$ is the convolutional displacement head, and
$I_i=I(\mathbf{x}_i^r)$ is the importance value at that node. The quantity
$w_i$ is its positive mean-normalized importance weight, $I_{\min}>0$ prevents
complete suppression of a node, $s_I$ controls the modulation range, $k$
indexes the $N$ nodes, and $\epsilon>0$ ensures numerical stability.
Finally, $\delta_{\max}$ bounds the displacement scale and
$\Delta\mathbf{x}_i$ is the resulting displacement. The reference coordinate
$\mathbf{x}_i^r$ and the number of nodes $N$ have been defined in
\autoref{sec:problem_formulation}.

The resulting path
$Q_g,Q_{\Delta},Q_E,Q_{\omega}\rightarrow I(\mathbf{x})
\rightarrow w_i\rightarrow\Delta\mathbf{x}_i$
explicitly connects physical structure to node relocation, thereby supporting
allocation readability. The nonuniform local discretizations produced by this
allocation motivate the interaction mechanism introduced next.

\subsection{Gauge-Aware Mesh Operator}
\label{sec:gauge_operator}

Adaptive allocation changes local spacing, orientation, neighborhood geometry,
and mesh Jacobians. GA-AMNO therefore applies an edge-conditioned feature
transport before aggregating information from neighboring adaptive nodes. For
each directed edge $j\rightarrow i$, the edge descriptor is
\begin{equation}
    \mathbf{e}_{ij}
    =
    \operatorname{Concat}\left(
    \mathbf{x}^{a}_{j}-\mathbf{x}^{a}_{i},
    \operatorname{vec}(\mathbf{J}^{a}_{j}),
    \operatorname{vec}(\mathbf{J}^{a}_{i}),
    \mathbf{p}_{j},
    \mathbf{p}_{i}
    \right).
    \label{eq:gauge_edge_descriptor}
\end{equation}
Here, $\mathbf{e}_{ij}$ is the descriptor of the edge from source node $j$ to
target node $i$, and $\mathbf{x}^{a}_{j}-\mathbf{x}^{a}_{i}$ is their relative
adaptive coordinate. The matrices $\mathbf{J}^{a}_{j}$ and
$\mathbf{J}^{a}_{i}$ are the estimated local mesh Jacobians at the source and
target nodes, respectively, $\operatorname{vec}(\cdot)$ flattens a matrix into
a vector, and $\mathbf{p}_{j}$ and $\mathbf{p}_{i}$ are their learned state
descriptors. The adaptive coordinates $\mathbf{x}^{a}_{i}$ have been defined
in \autoref{sec:problem_formulation}.

The transport uses a diagonal map with a gated low-rank correction:
\begin{equation}
    \mathbf{T}_{i\leftarrow j}^{(\ell)}
    =
    \operatorname{diag}(\mathbf{d}_{ij}^{(\ell)})
    +
    \frac{\gamma_{ij}^{(\ell)}}{\sqrt{C_h r}}
    \mathbf{U}_{ij}^{(\ell)}
    (\mathbf{V}_{ij}^{(\ell)})^{\top}.
    \label{eq:low_rank_transport}
\end{equation}
Here, $\mathbf{T}_{i\leftarrow j}^{(\ell)}
\in\mathbb{R}^{C_h\times C_h}$ is the source-to-target transport matrix in
layer $\ell$, $C_h$ is the hidden feature width, and $r$ is the transport
rank. The vector $\mathbf{d}_{ij}^{(\ell)}\in\mathbb{R}^{C_h}$ determines
diagonal channel scaling,
$\mathbf{U}_{ij}^{(\ell)},\mathbf{V}_{ij}^{(\ell)}
\in\mathbb{R}^{C_h\times r}$ determine the low-rank correction, and
$\gamma_{ij}^{(\ell)}$ is an edge-conditioned sigmoid gate. The superscript
$\top$ denotes matrix transpose. These quantities are predicted from
$\mathbf{e}_{ij}$ by separate learned heads.

Each source feature is transported into its target-conditioned representation
before weighted aggregation:
\begin{align}
    \alpha_{ij}^{(\ell)}
    &=
    \frac{\exp(a_{ij}^{(\ell)})}
    {\sum_{k\in\mathcal{N}(i)}\exp(a_{ik}^{(\ell)})},
    \nonumber\\
    \mathbf{m}_{i}^{(\ell)}
    &=
    \sum_{j\in\mathcal{N}(i)}
    \alpha_{ij}^{(\ell)}
    \mathbf{T}_{i\leftarrow j}^{(\ell)}
    \mathbf{h}_{j}^{(\ell)}.
    \label{eq:transported_aggregation}
\end{align}
Here, $a_{ij}^{(\ell)}$ is the learned compatibility score of edge
$j\rightarrow i$, and $\alpha_{ij}^{(\ell)}$ is its normalized attention
weight. The set $\mathcal{N}(i)$ contains the neighbors of target node $i$,
while $k$ indexes these neighbors in the softmax denominator.
The vector $\mathbf{h}_{j}^{(\ell)}$ is the source-node feature in layer
$\ell$, and $\mathbf{m}_{i}^{(\ell)}$ is the transported message aggregated
at target node $i$.

The target feature is then updated through
\begin{equation}
    \mathbf{h}_{i}^{(\ell+1)}
    =
    \mathbf{h}_{i}^{(\ell)}
    +
    \rho_{\theta}^{(\ell)}
    \left(\mathbf{m}_{i}^{(\ell)}\right).
    \label{eq:gauge_residual_update}
\end{equation}
Here, $\mathbf{h}_{i}^{(\ell)}$ and $\mathbf{h}_{i}^{(\ell+1)}$ are the
target-node features before and after the update, respectively, and
$\rho_{\theta}^{(\ell)}$ is the layer-specific update network. The residual
form preserves the current target feature while incorporating transported
neighborhood information.

The transport is interpreted as a learned, geometry-conditioned feature
correction rather than an analytically prescribed coordinate transformation.
Its representation-level motivation and complete implementation details are
provided in Appendix~\ref{app:gauge_operator_details}, while its theoretical
analysis is given in Appendix~\ref{sec:operator_readability_theory}. The
subsequent residual prediction is described
in Appendix~\ref{sec:pde_aware_correction}.

\section{Experimental Results}
\label{sec:experimental_results}

\subsection{Overall Predictive Performance}

\autoref{tab:main_results_revised} compares GA-AMNO with seven representative baselines covering spectral operators, graph operators, convolutional architectures, transformer-based operators, and adaptive-mesh solvers. GA-AMNO achieves the lowest central relative $\ell_2$ error on all five benchmarks. This consistent advantage across substantially different computational paradigms demonstrates that the proposed architecture is not specialized to outperform only one particular operator family. Instead, physics-informed adaptive allocation, low-rank gauge transport, and residual field refinement form an effective operator-learning framework that improves solution approximation across regular-grid, graph-based, and adaptive-mesh alternatives.

\begin{table*}[t]
\centering
\caption{Relative $\ell_2$ error (lower is better). All the models report mean $\pm$ standard deviation over three seeds. Classic denotes adaptive mesh learning with direct aggregation and the shared base solver. Best results are bold.}
\label{tab:main_results_revised}
\small
\setlength{\tabcolsep}{3pt}
\renewcommand{\arraystretch}{1.12}

\resizebox{\textwidth}{!}{%
\begin{tabular}{lcccccccc}
\hline
Dataset
& GA-AMNO
& Classic
& MMPDE
& Transolver
& FNO
& GNO
& WNO
& UNet \\
\hline

Navier-Stokes
& $\mathbf{0.003794 \pm 0.000624}$
& $0.011175 \pm 0.000065$
& $0.005300 \pm 0.000535$
& 0.019768
& 0.006781
& 0.033800
& 0.016903
& 0.005054 \\

Rayleigh-B\'enard
& $\mathbf{0.000927 \pm 0.000363}$
& $0.001472 \pm 0.000213$
& $0.003032 \pm 0.000477$
& 0.002664
& 0.006327
& 0.003954
& 0.007910
& 0.010840 \\

KS
& $\mathbf{0.000380 \pm 0.000061}$
& $0.000891 \pm 0.000128$
& $0.003988 \pm 0.001184$
& 0.013923
& 0.009289
& 0.010016
& 0.009549
& 0.010877 \\

Kolmogorov
& $\mathbf{0.020221 \pm 0.001529}$
& $0.081671 \pm 0.001860$
& $0.069614 \pm 0.003031$
& 0.070090
& 0.030034
& 0.090519
& 0.121220
& 0.028476 \\

Darcy
& $\mathbf{0.028157 \pm 0.001701}$
& $0.341614 \pm 0.004773$
& $0.030082 \pm 0.001907$
& 0.046329
& 0.028557
& 0.254866
& 0.118636
& 0.030238 \\

\hline
\end{tabular}%
}
\end{table*}

The comparisons with Classic and MMPDE are particularly important to the central claim of this work. Classic uses an adaptive mesh with direct feature aggregation, while MMPDE introduces a dedicated moving-mesh mechanism. GA-AMNO consistently outperforms both, showing that determining where nodes should be allocated is only part of successful adaptive operator learning. Explicitly transporting source features into the target representation context before aggregation provides an additional and practically useful capability for processing heterogeneous local discretizations. The strong results across Navier-Stokes, Rayleigh-B\'enard, KS, Kolmogorov, and Darcy further demonstrate that this benefit extends from evolutionary and chaotic dynamics to static coefficient-to-solution mappings. Overall, the main experiment supports the central design principle of operator readability: the predictive benefit emerges when adaptive discretization is operator-readable, while broad applicability is retained across distinct PDE regimes.

\subsection{Component Ablation}
\label{sec:component_ablation}

\begin{figure*}[t]
\centering
\includegraphics[width=0.92\textwidth]{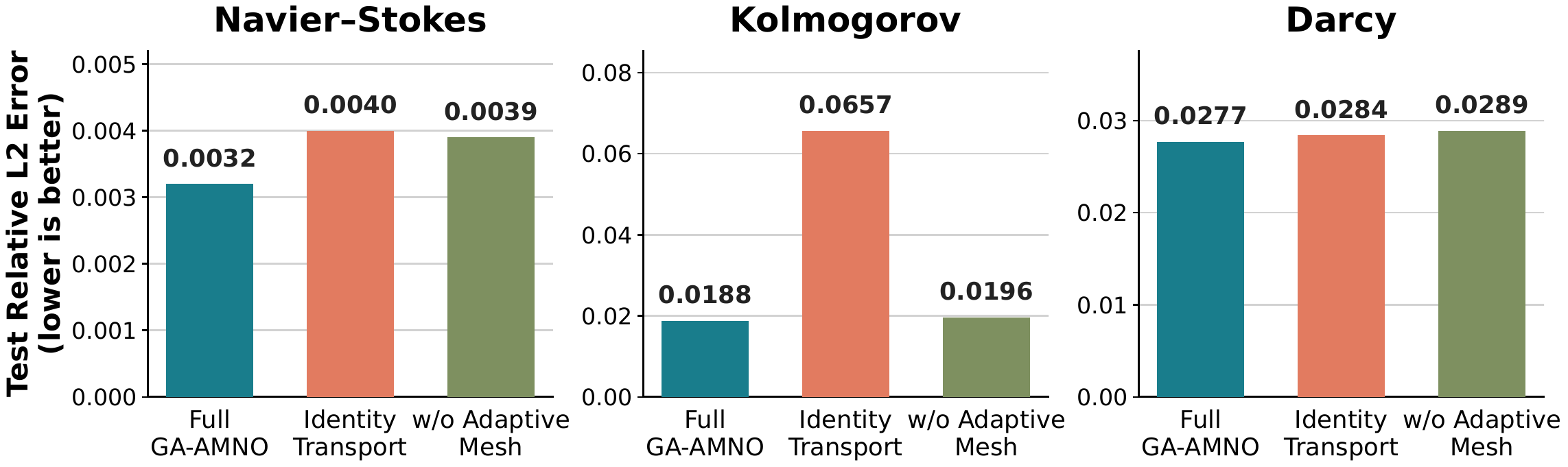}
\caption{Ablation of the two core mechanisms. Identity Transport retains the adaptive mesh but removes learned representation transport, while w/o Adaptive Mesh retains the remaining solver without learned node relocation. Lower test relative $\ell_2$ error is better.}
\label{fig:strict_mechanism_ablation}
\end{figure*}

We isolate the two mechanisms central to operator-readable adaptation while retaining the remaining architecture and evaluation protocol. Identity Transport replaces every learned source-to-target transport with the identity map, whereas w/o Adaptive Mesh removes the learned node relocation. \autoref{fig:strict_mechanism_ablation} reports the resulting test relative $\ell_2$ errors on Navier-Stokes, Kolmogorov, and Darcy.

Full GA-AMNO achieves the lowest error on all three datasets. Replacing learned transport with the identity increases error, most substantially on Kolmogorov, while removing adaptive allocation also consistently degrades performance. The comparison directly verifies that both where representations are allocated and how they are aligned contribute to the complete model. Targeted ablations of the remaining components are provided in \autoref{fig:targeted_module_ablation} in the extended results.

\subsection{Interpreting Adaptive Computation}
\label{sec:interpreting_adaptive_computation}

This subsection examines whether the learned importance map is used by the predictor, whether the learned connection identifies influential edges, and whether transport converts source messages into target-compatible representations under strong local geometric mismatch. All experiments use frozen checkpoints and held-out samples, with no parameter updates during evaluation. Prediction effects in high-adaptation-demand regions are reported in the targeted component ablation of \autoref{fig:targeted_module_ablation}, avoiding duplicate tabulation of the same quantities.

\subsubsection{Importance Allocation Faithfulness}

We intervene on the learned importance map while leaving the remaining network unchanged. The learned map is replaced by a spatially uniform map, randomly shuffled within each sample, or cyclically shifted. If the map merely visualized activity without controlling useful computation, these counterfactuals would leave the prediction nearly unchanged.

As shown in \autoref{fig:causal_interventions}(a), the learned map is best in all three cases, but the impact varies across PDEs. Navier-Stokes exhibits a strong counterfactual gap, Darcy exhibits a moderate gap, and Kolmogorov is only weakly sensitive. The counterfactual design is important because it distinguishes a visually plausible importance map from one that actually controls prediction: preserving its values while destroying their spatial correspondence degrades performance. This experiment therefore supports allocation readability by showing that the learned spatial allocation is functionally used, although it does not prove that every highlighted region has a unique physical interpretation or that importance identifiability is equally strong for every PDE\@.

\subsubsection{Gauge Edge Faithfulness}

Using the correction magnitude and prediction-change metric defined in \autoref{sec:experimental_setting}, we replace the learned transport on a selected $10\%$ of edges by the identity map. Top-ranked, random, and bottom-ranked edge subsets are evaluated under the same intervention budget.

\begin{figure*}[t]
\centering
\includegraphics[width=0.94\textwidth]{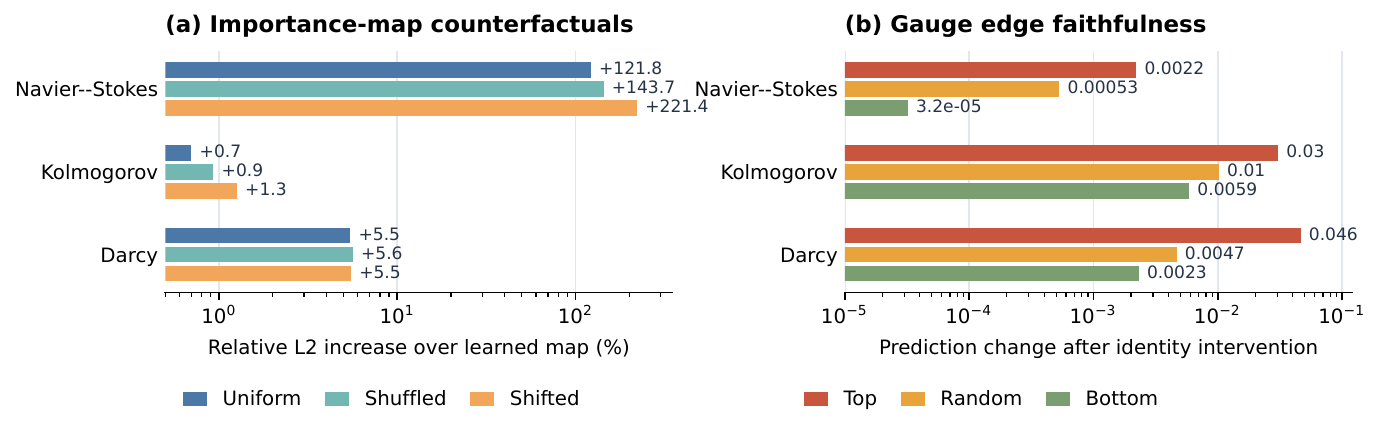}
\caption{Causal tests of allocation and interaction readability. (a) Relative $\ell_2$ increase after replacing the learned importance map with uniform, shuffled, or shifted alternatives. (b) Prediction change after replacing transport with the identity on top-ranked, random, or bottom-ranked edge subsets.}
\label{fig:causal_interventions}
\end{figure*}

As shown in \autoref{fig:causal_interventions}(b), top-ranked interventions produce $2.99$--$9.92$ times the prediction change of random interventions, while bottom-ranked edges have little influence. The correction magnitude is therefore functionally faithful to the trained operator: edges assigned a large representation repair are also the edges on which the output causally depends. The purpose of this intervention is to establish a causal link between an interpretable internal quantity and the final prediction, which cannot be obtained from feature visualization alone. It supports interaction readability by showing that the learned connection identifies consequential information transfers rather than acting as a decorative latent variable. The correction magnitude is not itself claimed to be a physical observable.

\subsubsection{Message Compatibility by Local Geometric Mismatch}

We use the geometric-mismatch groups and compatibility metrics defined in \autoref{sec:experimental_setting} to compare source--target discrepancy with transported-message--target discrepancy. This evaluation asks whether the learned correction is most active where the adaptive geometry is most nonuniform; it does not by itself separate physical feature variation from discretization-induced variation.

\begin{figure*}[t]
\centering
\includegraphics[width=0.94\textwidth]{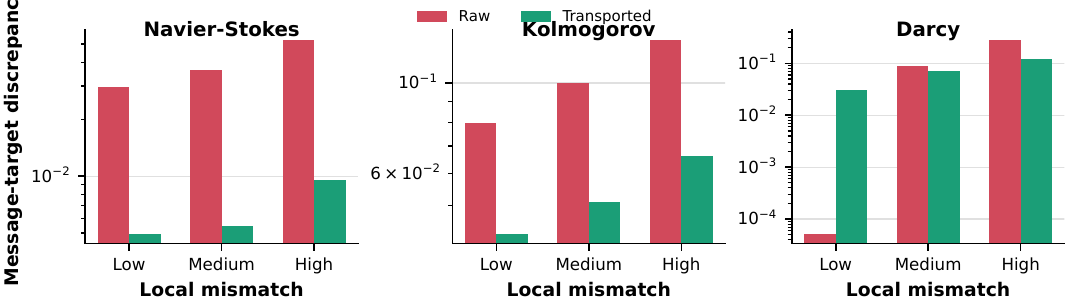}
\caption{Source--target and transported-message--target discrepancy stratified by local geometric mismatch. The reduction measures message compatibility before aggregation, not the covariance defect of the learned connection.}
\label{fig:feature_alignment_revised}
\end{figure*}

As summarized in \autoref{tab:feature_alignment_revised}, transport reduces message--target discrepancy by $48.32\%$--$81.93\%$ in the high-geometric-mismatch group. The full stratification in \autoref{fig:feature_alignment_revised} shows that this compatibility gain is concentrated where the adaptive geometry is most nonuniform. This concentration is the key mechanism-level result: the learned correction is strongest where direct comparison is least justified by the local discretization, which is the regime targeted by interaction readability. Together with the edge intervention and regional prediction ablation, the result connects geometric mismatch, feature correction, and output utility. Because the metric also contains actual source--target physical variation, it remains a compatibility diagnostic rather than a standalone proof of gauge covariance.

\section{Conclusion}
We introduced GA-AMNO to address a structural issue in adaptive neural operators: state-dependent discretization changes not only where information is sampled, but also the local representation context in which information is encoded and exchanged. GA-AMNO therefore treats adaptive computation as two coupled problems, using physics-informed adaptive allocation to determine where resolution should be allocated and low-rank gauge transport to determine how features associated with unequal local discretizations should interact before aggregation. The theoretical analysis explains why shared direct aggregation cannot generally remove discretization-induced representation ambiguity, establishes sufficient conditions for consistent transported aggregation, and bounds the effect of approximate learned connections without claiming exact gauge equivariance. The counterfactual, intervention, mismatch-stratified, and deformation experiments further show that the learned allocation and transport are functionally used rather than merely producing visually plausible meshes or latent variables. The broader significance of this work is to reposition adaptive discretization from an opaque preprocessing or resource-allocation mechanism into a mechanism that operationalizes operator readability, with allocation decisions and information transfers that can both be inspected and tested. Future work should extend this formulation to explicitly equivariant connection parameterizations, topology-changing and anisotropic mesh adaptation, three-dimensional unstructured domains, and scalable sparse transport, while developing stronger links between representation consistency, numerical approximation error, and the convergence properties of learned PDE operators.

\bibliography{iclr2027_conference}
\bibliographystyle{iclr2027_conference}

\appendix
\raggedbottom

% Permit appendix floats to share pages with surrounding text instead of
% being deferred to sparsely populated float-only pages.
\setcounter{topnumber}{5}
\setcounter{dbltopnumber}{3}
\renewcommand{\topfraction}{0.95}
\renewcommand{\dbltopfraction}{0.95}
\renewcommand{\textfraction}{0.05}
\renewcommand{\floatpagefraction}{0.85}
\renewcommand{\dblfloatpagefraction}{0.85}

\section{Algorithmic Summary}
\label{app:algorithm}

\autoref{tab:ga_amno_algorithm} summarizes the forward procedure of
GA-AMNO\@. All learnable components are optimized jointly during training.

\begin{table*}[!t]
\centering
\caption{Forward procedure of the proposed GA-AMNO\@.}
\label{tab:ga_amno_algorithm}
\small
\renewcommand{\arraystretch}{1.12}
\setlength{\tabcolsep}{6pt}
\begin{tabular}{p{0.08\textwidth}p{0.84\textwidth}}
\hline
\textbf{Step} & \textbf{Operation} \\
\hline

\textbf{Input}
&
Receive the temporal history or steady conditioning field $\mathbf{a}$
on the reference discretization $\mathcal{X}^r$.
\\

\textbf{1}
&
Encode the input into a state descriptor:
$\mathbf{P}=\operatorname{Encoder}(\mathbf{a})$.
\\

\textbf{2}
&
Compute gradient, Laplacian, energy, and vorticity-related indicators
from the latest or conditioning field $\mathbf{v}$.
\\

\textbf{3}
&
Predict the importance map from the encoded state and normalized
physical indicators:
$I=\sigma\!\left(\eta_\theta(\operatorname{Concat}
(\mathbf{P},\widetilde{\mathbf{Q}}))\right)$.
\\

\textbf{4}
&
Predict bounded node directions and construct importance-modulated
adaptive coordinates:
$\mathbf{x}_i^a=\Pi_\Omega
(\mathbf{x}_i^r+\delta_{\max}w_i\mathbf{o}_i)$.
\\

\textbf{5}
&
Estimate local mesh Jacobians $\mathbf{J}_i^a$ and form each directed
edge descriptor $\mathbf{e}_{ij}$ from relative coordinates, source and
target Jacobians, and source and target state descriptors.
\\

\textbf{6}
&
Predict the edge attention score and diagonal-plus-low-rank transport:
$\mathbf{T}_{i\leftarrow j}^{(\ell)}
=\operatorname{diag}(\mathbf{d}_{ij}^{(\ell)})
+\gamma_{ij}^{(\ell)}
\mathbf{U}_{ij}^{(\ell)}
(\mathbf{V}_{ij}^{(\ell)})^\top/\sqrt{C_hr}$.
\\

\textbf{7}
&
Transport every source feature into its target-conditioned
representation:
$\widetilde{\mathbf{h}}_{i\leftarrow j}^{(\ell)}
=\mathbf{T}_{i\leftarrow j}^{(\ell)}
\mathbf{h}_j^{(\ell)}$.
\\

\textbf{8}
&
Aggregate transported messages and update the target feature:
$\mathbf{m}_i^{(\ell)}
=\sum_{j\in\mathcal{N}(i)}
\alpha_{ij}^{(\ell)}
\widetilde{\mathbf{h}}_{i\leftarrow j}^{(\ell)}$ and
$\mathbf{h}_i^{(\ell+1)}
=\mathbf{h}_i^{(\ell)}
+\rho_\theta^{(\ell)}(\mathbf{m}_i^{(\ell)})$.
\\

\textbf{9}
&
Repeat Steps 6--8 for all $L_g$ low-rank gauge transport layers and
modulate the final nodal features by the learned importance weights.
\\

\textbf{10}
&
Reconstruct the adaptive nodal features on the canonical grid using
coarse-grid upsampling or distance-weighted interpolation, producing
$\mathbf{F}_{\mathrm{grid}}$.
\\

\textbf{11}
&
Predict the local residual
$\boldsymbol{\delta}_{\mathrm{raw}}$ and the truncated spectral correction
$\boldsymbol{\delta}_{\mathrm{spec}}$, and set
$\boldsymbol{\delta}_1=
\boldsymbol{\delta}_{\mathrm{raw}}+
\boldsymbol{\delta}_{\mathrm{spec}}$.
\\

\textbf{12}
&
Construct the provisional field $\mathbf{u}_{\mathrm{base}}$, compute its
finite-difference features, and obtain the differential correction and
final residual $\boldsymbol{\delta}$.
\\

\textbf{Output}
&
Return
$\widehat{\mathbf{u}}=\mathbf{u}_t+\boldsymbol{\delta}$
for temporal residual prediction, or
$\widehat{\mathbf{u}}=\boldsymbol{\delta}$
for steady or direct-output problems.
\\

\hline
\end{tabular}
\end{table*}

\section{Expanded Related Work}
\label{sec:expanded_related_work}

\subsection{Neural Operators for PDE Solving}

Neural operators learn mappings between function spaces and provide a practical route to surrogate PDE solvers that can share parameters across discretization resolutions \cite{huang2025pde,kovachki2023neuraloperator}. Deep Operator Network (DeepONet) represents an input function and its query coordinates through branch and trunk networks \cite{lu2021deeponet}, whereas Fourier Neural Operator (FNO) parameterizes global integral operators in the spectral domain \cite{li2020fourier}. This basic formulation has subsequently been extended through factorized spectral layers \cite{tran2023factorized}, adaptive Fourier token mixing \cite{guibas2022afno}, U-shaped multiscale architectures \cite{rahman2023uno}, wavelet decompositions \cite{tripura2023wno}, and physics-informed objectives \cite{li2024pino}. Graph Neural Operator (GNO) \cite{li2020neural} and Message Passing Neural PDE Solvers \cite{brandstetter2022message} support local interactions on irregular samples, while Geo-FNO \cite{li2023geofno} learns a deformation between irregular physical domains and a regular latent grid. Geometry-Informed Neural Operator further combines graph-based lifting with spectral processing for large-scale PDEs on varying geometries \cite{li2023geometry}. Transformer-based operators, including GNOT \cite{hao2023gnot} and Transolver \cite{wu2024transolver}, model long-range physical interactions, while explicit space--frequency coupling improves the recovery of multiscale solution features \cite{bie2025space}. More recent work has addressed stable long-horizon prediction through iterative spectral refinement \cite{lippe2023pde}, general-purpose PDE pretraining through the Poseidon foundation model \cite{herde2024poseidon}, and the integration of global spectra with local derivative information through the Riesz Neural Operator \cite{liu2026riesz}. Despite these advances, the numerical realization of a neural operator still acts on discretized field representations. Representation Equivalent Neural Operators show that changes in sampling and reconstruction can introduce operator aliasing and break consistency between continuous operators and their discrete realizations \cite{bartolucci2023representation}. This observation is closely related to our motivation, but our focus is different: rather than addressing global continuous--discrete equivalence alone, we study how features produced under unequal, input-dependent local discretizations should interact inside an adaptive-mesh operator.

\subsection{Data-driven Adaptive Mesh Methods}

Adaptive meshing concentrates computational resolution near sharp gradients, discontinuities, and dynamically complex structures. Classical adaptive mesh refinement uses equation-dependent error indicators and hierarchical refinement rules \cite{berger1984adaptive,berger1989local}, while constrained neural adaptation provides an early example of learning spatial resolution for PDE approximation \cite{rudd2014adaptive}. Graph Element Networks introduce adaptive structured computation over spatial elements \cite{alet2019graph}, and graph-based simulators subsequently support physical prediction on mesh and particle representations \cite{sanchezgonzalez2020simulate}. MeshGraphNets learn dynamics directly on simulation meshes and adapt connectivity during rollout \cite{pfaff2020learning}, whereas differentiable combinations of PDE solvers and graph networks preserve a stronger connection to conventional numerical computation \cite{belbuteperes2020differentiable}. More explicit learned-mesh approaches treat node placement itself as an optimization problem. M2N learns end-to-end mesh movement for PDE solvers \cite{song2022m2n}, and LAMP jointly learns physical evolution with a controllable refinement and coarsening policy for multiresolution simulation \cite{wu2023learning}. Data-Free Mesh Movers construct an $r$-adaptive mesh through a Monge--Amp\`ere-based objective and integrate the resulting mesh into a learned PDE solver \cite{hu2024better}. UGM2N further introduces an unsupervised mesh-movement objective based on local equidistribution \cite{wang2025ugm2n}. Related operator-learning methods handle geometric variation through learned global coordinate transformations: Geo-FNO maps irregular geometries to a uniform latent domain \cite{li2023geofno}, while GINO transfers information between irregular point sets and a regular latent grid \cite{li2023geometry}. These methods substantially improve where computation is allocated or how irregular data are embedded. However, the adapted mesh is still commonly treated as a sampling, interpolation, or message-passing structure once it has been generated. Node relocation changes local density, orientation, spatial support, and neighborhood geometry, but conventional aggregation does not explicitly account for how these changes affect the representation convention of each node feature. Consequently, physical variation and discretization-induced representation variation may be mixed during information exchange. GA-AMNO addresses this complementary problem by coupling physics-informed adaptive allocation with source-to-target feature transport before aggregation.

\subsection{Equivariant and Gauge-Aware Representation Learning}

Equivariant representation learning provides a principled framework for processing features under transformations of coordinates, orientations, or local reference frames. Geometric deep learning unifies many of these constructions across grids, groups, graphs, and manifolds \cite{bronstein2021geometric}. Steerable three-dimensional convolutions \cite{weiler20183d}, $\mathrm{SE}(3)$-Transformers \cite{fuchs2020se3}, and $E(n)$-equivariant graph networks \cite{satorras2021egnn} preserve prescribed transformation laws under global rotations and translations. In operator learning, Group Equivariant Fourier Neural Operators extend rotational, translational, and reflectional symmetries to spectral operator layers \cite{helwig2023group}. Steerable partial differential operators provide a complementary characterization of when differential mappings between feature fields are equivariant \cite{jenner2021steerable}. These global symmetry constructions improve physical consistency, but they do not directly resolve the comparison of features expressed in independently varying local frames. Gauge-equivariant convolutional networks instead associate each location with a local representation frame and require feature transformations to follow a prescribed gauge action \cite{cohen2019gauge}. Gauge Equivariant Mesh CNNs extend this principle to manifold meshes by parallel-transporting neighboring features before applying anisotropic convolution \cite{dehaan2021gauge}. Gauge Equivariant Transformer incorporates parallel transport into self-attention on triangular meshes \cite{he2021gauge}, while gauge-equivariant nonlinear message passing applies the same geometric principle to nonlinear dynamics and PDE evolution on surfaces \cite{park2023modeling}. These studies establish the central principle that features expressed under different local conventions should be transported into compatible representations before interaction. However, they generally assume a fixed geometric domain, explicitly constructed local frames, and predefined transformation laws. GA-AMNO considers a different setting in which the discretization itself depends on the input. The resulting local representation changes are not known a priori and are not claimed to obey exact gauge equivariance. Instead, relative coordinates, adaptive-mesh Jacobians, and state descriptors condition a learned diagonal-plus-low-rank transport that maps each source feature into a target-conditioned representation before aggregation. The gauge interpretation therefore provides a model for interaction readability under input-dependent discretization, while the learned connection is treated as an approximate, functionally testable representation-alignment mechanism.

\section{Experimental Setting}
\label{sec:experimental_setting}

\subsection{Datasets and Data Preparation}

We evaluate GA-AMNO on five PDE benchmarks: Navier-Stokes \cite{li2025latent}, Rayleigh-B\'enard \cite{burns2020dedalus}, KS \cite{shysheya2024conditional}, Kolmogorov \cite{rozet2023score}, and Darcy \cite{li2025latent}. The benchmarks cover scalar and multi-channel temporal prediction as well as a steady-state coefficient-to-solution mapping.

The Navier-Stokes dataset contains $5{,}000$ trajectories of two-dimensional incompressible flow at viscosity $10^{-3}$. Each trajectory contains $50$ scalar flow-state snapshots on a $64\times64$ grid. We use trajectories $1$--$1{,}000$ for training, $1{,}001$--$1{,}100$ for validation, and the final $100$ trajectories for testing. The Rayleigh-B\'enard dataset contains vorticity and temperature-anomaly fields on a $64\times64$ grid. Its training, validation, and test splits contain $256$, $64$, and $128$ trajectories, respectively, with $41$ snapshots per trajectory. The KS dataset contains one-dimensional scalar trajectories with $256$ spatial points and $641$ temporal snapshots. The training, validation, and test splits contain $800$, $100$, and $50$ trajectories, respectively. The Kolmogorov dataset contains two-channel forced-flow states on a $64\times64$ grid. It provides $819$ training trajectories, $102$ validation trajectories, and $103$ test trajectories, with $64$ snapshots per trajectory. The Darcy benchmark is a steady-state operator-learning problem on a $64\times64$ grid. Its training, validation, and test splits contain $900$, $124$, and $1{,}024$ samples, respectively. Each sample stores a coefficient field followed by its normalized solution field.

For the four temporal benchmarks, the input is a history of four consecutive snapshots, $\mathcal{U}_{t}=\{u_{t-3},u_{t-2},u_{t-1},u_t\}$, and the target is the immediately following state $u_{t+1}$. The starting indices of successive windows are separated by $8$ snapshots for Navier-Stokes, Rayleigh-B\'enard, and KS, and by $4$ snapshots for Kolmogorov. Thus, the stride controls window extraction and does not introduce gaps between the four states inside a history window. For Darcy, the coefficient field is used as a single conditioning input and the corresponding solution is the target. No additional dataset-level normalization is applied during training; the pre-normalized Darcy files are loaded directly.

\subsection{Baselines and Comparison Protocol}

We compare GA-AMNO with seven representative baselines. 
\textbf{1)} FNO \cite{li2020fourier} applies global spectral convolution on a regular grid. 
\textbf{2)} Transolver \cite{wu2024transolver} uses physics-aware slice attention to model long-range spatial interactions. 
\textbf{3)} GNO \cite{li2020neural} performs coordinate-conditioned local message passing. 
\textbf{4)} WNO \cite{tripura2023wno} uses Haar-wavelet operator blocks to capture localized multiresolution components. 
\textbf{5)} UNet \cite{ronneberger2015u} is a convolutional encoder--decoder with multiscale skip connections. 
\textbf{6)} Classic is a controlled adaptive-mesh baseline that learns node displacements but directly aggregates node features using the shared base solver, without physics-informed adaptive allocation, low-rank gauge transport, spectral residual correction, or PDE-aware correction. 
\textbf{7)} MMPDE \cite{hu2024better} combines a data-free differentiable mesh mover with regular-mesh and moving-mesh solution branches to perform adaptive PDE prediction. 
Our implementations of Transolver, GNO, WNO, and MMPDE follow the core mechanisms of the cited architectures and are adapted to the unified input--output protocol, spatial resolution, and training configuration used in this study. Classic shares the corresponding base prediction protocol with GA-AMNO and serves as a controlled comparison for evaluating the benefit of operator-readable adaptive discretization beyond adaptive node movement alone.

All methods use the same dataset files, trajectory splits, history windows, prediction targets, spatial resolutions, learning rate, epoch count, and number of updates per epoch. Models that use coordinates receive the same regular physical-coordinate encoding. Checkpoints are selected by the same validation relative $\ell_2$ criterion. The comparison controls the data protocol and optimization-step budget but does not enforce parameter- or FLOP-matched architectures. Parameter counts and counted FLOPs are therefore reported separately when computational cost is discussed.

\subsection{Evaluation Metrics}

We use relative $\ell_2$ error as the primary metric for predictive comparison. RMSE and spectral-energy error are additionally reported for rollout evaluation, while gradient, differential, and high-band spectral errors are used for targeted mechanism analyses.
\begin{align}
\mathrm{RelL2}(\widehat{u},u)&=\frac{\|\widehat{u}-u\|_2}{\|u\|_2+\epsilon},\nonumber\\
\mathrm{RMSE}(\widehat{u},u)&=\sqrt{\frac{1}{N}\sum_{i=1}^{N}(\widehat{u}_i-u_i)^2},\nonumber\\
\mathrm{SpectrumError}(\widehat{u},u)&=\frac{\|\overline{|\mathcal{F}_{\mathrm{DFT}}(\widehat{u})|^2}_{c}-\overline{|\mathcal{F}_{\mathrm{DFT}}(u)|^2}_{c}\|_2}{\|\overline{|\mathcal{F}_{\mathrm{DFT}}(u)|^2}_{c}\|_2+\epsilon}.
\end{align}
Here, $\widehat{u}$ and $u$ are the prediction and ground truth, $N$ is the number of scalar field entries, $\mathcal{F}_{\mathrm{DFT}}$ is the orthonormal discrete Fourier transform, $\overline{(\cdot)}_{c}$ denotes averaging over channels, and $\epsilon=10^{-8}$. The overall $\mathrm{SpectrumError}$ measures spectral-energy discrepancy over all represented frequencies and is used for rollout evaluation. For the Spectral Residual targeted ablation, we separately report high-band spectral error. Let
\begin{equation*}
    \mathcal{K}_{\mathrm{hi}}=\left\{k:\frac{\nu(k)}{\nu_{\max}}\geq 0.6\right\},
\end{equation*}
where $\nu(k)$ is the radial frequency of coefficient $k$ and $\nu_{\max}$ is the maximum represented radius. The high-band spectral error is the same channel-averaged spectral-energy discrepancy restricted to $\mathcal{K}_{\mathrm{hi}}$:
\begin{equation*}
\mathrm{HighBandError}(\widehat{u},u)=\frac{\left\|\overline{|\mathcal{F}_{\mathrm{DFT}}(\widehat{u})|^2}_{c,\mathcal{K}_{\mathrm{hi}}}-\overline{|\mathcal{F}_{\mathrm{DFT}}(u)|^2}_{c,\mathcal{K}_{\mathrm{hi}}}\right\|_2}{\left\|\overline{|\mathcal{F}_{\mathrm{DFT}}(u)|^2}_{c,\mathcal{K}_{\mathrm{hi}}}\right\|_2+\epsilon}.
\end{equation*}
Here, the subscript $c,\mathcal{K}_{\mathrm{hi}}$ denotes channel averaging followed by restriction to the high-band coefficients. Gradient error is used in mechanism and regional analyses:
\begin{equation}
\mathrm{GradError}(\widehat{u},u)=\frac{\|[D_x\widehat{u}-D_xu,D_y\widehat{u}-D_yu]\|_2}{\|[D_xu,D_yu]\|_2+\epsilon},
\end{equation}
where $D_x$ and $D_y$ are centered finite differences in grid-index units with one-sided boundary differences. For one-dimensional KS data, the inactive vertical derivative is zero.

For Darcy, the current auxiliary differential diagnostic is reported as Laplacian discrepancy,
\begin{equation}
E_{\Delta}=\frac{\|\Delta\widehat{u}-\Delta u\|_2}{\|\Delta u\|_2+\epsilon},
\end{equation}
rather than as an equation residual, because it does not explicitly evaluate $-\nabla\cdot(a\nabla\widehat{u})-f$. For scalar Navier-Stokes and KS fields, divergence, vector vorticity, and enstrophy are not reported because they are not defined by the stored single-channel representation. Physics-specific vector-field diagnostics are used only when the stored channels have the required physical meaning.

To study input-dependent adaptation demand, we use
\begin{align}
d(\mathbf{x})={}&0.4\mathcal{N}(g(\mathbf{x}))+0.3\mathcal{N}(c(\mathbf{x}))\nonumber\\
&+0.2\mathcal{N}(m(\mathbf{x}))+0.1\mathcal{N}(I(\mathbf{x})),
\end{align}
where $g=\|\nabla u_t\|$ is input-gradient magnitude, $c=|\Delta u_t|$ is input-curvature magnitude, $m$ is the norm of the local finite-difference variation of the learned mesh offset, and $I$ is the learned importance map. $\mathcal{N}$ denotes per-sample robust normalization using the $5$th and $95$th percentiles. The lower, middle, and upper terciles of $d$ define low-, medium-, and high-demand masks. The masks are constructed once from the input and the full GA-AMNO geometry, without using targets or predictions, and the same masks are applied to every compared model.

Importance allocation faithfulness is evaluated by replacing the learned importance map with a spatially uniform, randomly shuffled, or cyclically shifted map and then reporting the resulting relative $\ell_2$ error. Gauge edge faithfulness is measured using the transport-correction magnitude
\begin{equation}
c_{ij}=\left\|\mathbf{T}_{i\leftarrow j}\mathbf{h}_{j}-\mathbf{h}_{j}\right\|_2,
\label{eq:edge_faithfulness_score_revised}
\end{equation}
where $\mathbf{h}_{j}$ is the source feature and $\mathbf{T}_{i\leftarrow j}$ is the learned transport from node $j$ to node $i$. After replacing the transport on an edge subset $S$ by the identity, its causal influence is quantified as
\begin{equation}
\Delta_{\mathrm{pred}}(S)=\frac{\|\widehat{u}_{S}-\widehat{u}_{\mathrm{full}}\|_2}{\|\widehat{u}_{\mathrm{full}}\|_2+\epsilon},
\label{eq:edge_prediction_change_revised}
\end{equation}
where $\widehat{u}_{S}$ is the intervened prediction and $\widehat{u}_{\mathrm{full}}$ is the original prediction.

Message--target compatibility is evaluated after dividing directed edges into low-, medium-, and high-geometric-mismatch groups using local edge-scale ratio, mesh-Jacobian variation, and importance difference. For each group $\mathcal{E}_{b}$, we compute
\begin{align}
E_{\mathrm{src}}^{(b)}&=\frac{1}{|\mathcal{E}_{b}|}\sum_{(i,j)\in\mathcal{E}_{b}}\frac{\|\mathbf{h}_{i}-\mathbf{h}_{j}\|_2}{\|\mathbf{h}_{i}\|_2+\epsilon},\nonumber\\
E_{\mathrm{msg}}^{(b)}&=\frac{1}{|\mathcal{E}_{b}|}\sum_{(i,j)\in\mathcal{E}_{b}}\frac{\|\mathbf{h}_{i}-\mathbf{T}_{i\leftarrow j}\mathbf{h}_{j}\|_2}{\|\mathbf{h}_{i}\|_2+\epsilon},\nonumber\\
G_{\mathrm{comp}}^{(b)}&=\frac{E_{\mathrm{src}}^{(b)}-E_{\mathrm{msg}}^{(b)}}{E_{\mathrm{src}}^{(b)}+\epsilon}.
\label{eq:alignment_metrics_revised}
\end{align}
Here, $b$ indexes the geometric-mismatch group, $E_{\mathrm{src}}^{(b)}$ is the source--target feature discrepancy, $E_{\mathrm{msg}}^{(b)}$ is the transported-message--target discrepancy, and $G_{\mathrm{comp}}^{(b)}$ is their relative reduction. A positive value means that the message supplied to aggregation is closer to the current target feature. Because neighboring nodes may contain actually different physical content, this quantity is a message-compatibility diagnostic rather than a direct estimate of the covariance defect in \autoref{eq:learned_consistency_defects}. The training-time loop diagnostic constrains only the diagonal transport component around sampled local triangular cycles and is therefore termed diagonal loop consistency, not exact holonomy of the complete low-rank transport. Mesh inversion rate, minimum cell angle, and aspect ratio are reported only as mesh-health diagnostics.

\subsection{Implementation Details}

GA-AMNO uses a hidden width of $32$, two low-rank gauge transport layers, six indexed neighbors per node, and a low-rank transport rank of $r=4$. The operator stride is $2$ for the two-dimensional datasets and $1$ for KS\@. The maximum latent-mesh displacement is $0.08$ for the temporal benchmarks and is reduced to $0.02$ for Darcy. Dataset-specific spectral modes and auxiliary-loss weights are selected using only the validation split. The reference connectivity is fixed, while latent coordinates, local Jacobians, edge descriptors, and transport matrices depend on the input state.

All methods are optimized with AdamW for $300$ fixed-budget epochs using an initial learning rate of $10^{-3}$. Each epoch consists of $100$ mini-batch updates, giving $30{,}000$ parameter updates per run rather than $300$ complete passes through each windowed dataset. The default training batch size is $8$. Owing to GPU-memory constraints, GA-AMNO uses a batch size of $4$ on Rayleigh-B\'enard, whereas the corresponding baseline runs use batch size $8$. Validation is performed every $10$ epochs, in addition to the first and final epochs, and the checkpoint with the lowest validation relative $\ell_2$ error is selected for testing. GA-AMNO is evaluated using three random seeds, $42$, $43$, and $44$, and its main results are reported as the mean and standard deviation across these runs. Unless otherwise specified, the baseline and mechanism-analysis experiments use random seed $42$.

The GA-AMNO objective combines field MSE with supervised gradient, Laplacian, and spectral losses, together with diagonal loop consistency, transport alignment, mesh-quality, physics-informed allocation, frame-consistency, cross-mesh, and mesh-perturbation regularizers. The weights are tuned separately for each dataset on its validation split and remain fixed during test evaluation. These auxiliary terms regularize the learned latent geometry and feature transport; they are not treated as exact gauge-equivariance constraints.

\subsection{Efficiency and Memory Scalability}

We evaluate computational scalability using the Navier-Stokes GA-AMNO checkpoint on an NVIDIA GeForce RTX 4070 Laptop GPU\@. The model contains $177.4$K trainable parameters. At $64\times64$ resolution with batch size $8$, the hook-based estimator reports $8.943$G counted FLOPs for convolutional and linear operations. The measured forward latency is $23.38$ ms per batch and peak allocated forward memory is $144.3$ MiB. Latency is measured after $15$ warm-up passes and averaged over $30$ synchronized inference passes. The FLOP estimate counts a multiply--add as two operations and does not include every FFT, distance, interpolation, normalization, softmax, or tensor-contraction operation; it is therefore reported as an implementation-level estimate rather than an exact hardware-independent operation count.

\begin{figure*}[!t]
\centering
\begin{minipage}[t]{0.485\textwidth}
\centering
\textbf{(a)}\par\vspace{2pt}
\includegraphics[width=\linewidth]{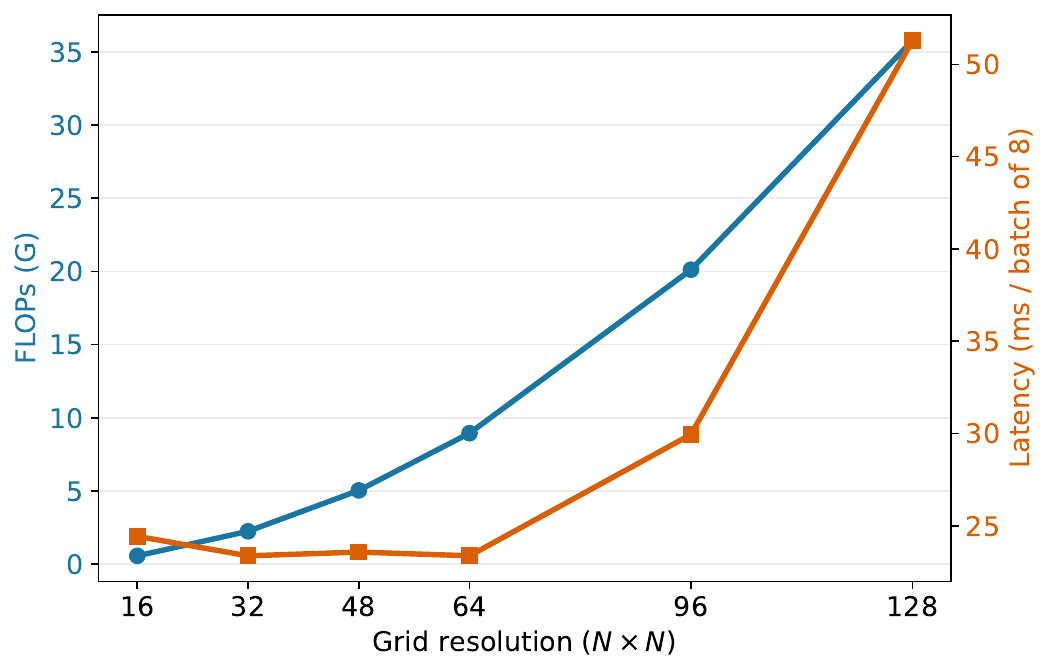}
\end{minipage}
\hfill
\begin{minipage}[t]{0.485\textwidth}
\centering
\textbf{(b)}\par\vspace{2pt}
\includegraphics[width=\linewidth]{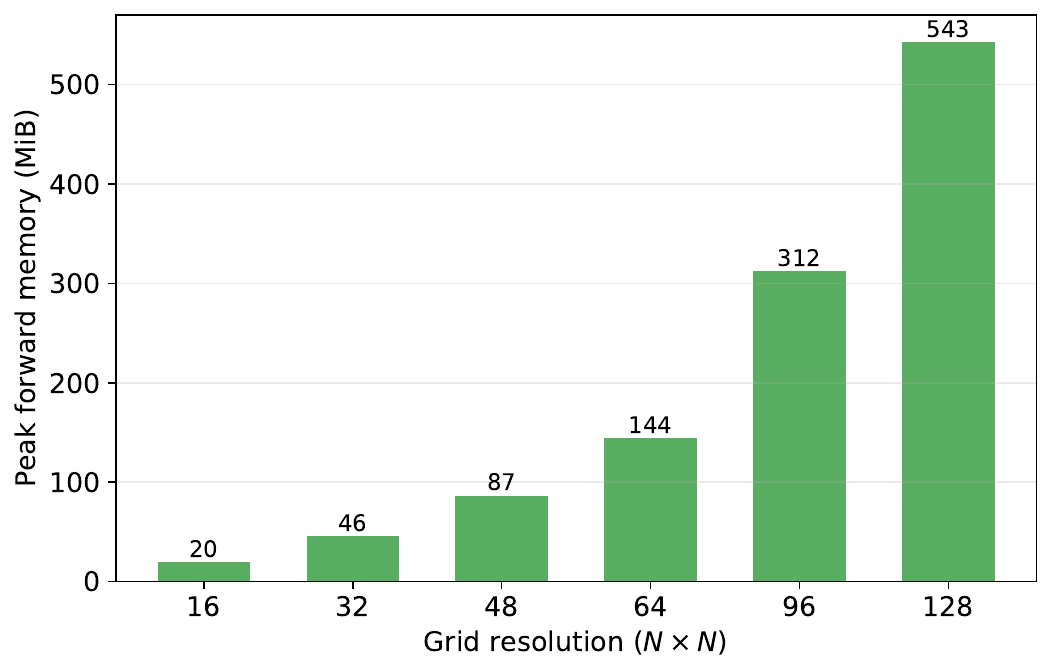}
\end{minipage}
\caption{Computational scalability of GA-AMNO\@. (a) Counted FLOPs and synchronized forward latency at different grid resolutions. (b) Peak allocated forward memory. The same Navier-Stokes checkpoint and batch size $8$ are used throughout.}
\label{fig:gauge_scalability}
\end{figure*}

As shown in \autoref{fig:gauge_scalability}, the same checkpoint is evaluated from $16\times16$ to $128\times128$ without changing model parameters. Inputs are bilinearly resized solely for this computational-scaling study. The counted FLOPs increase from $0.559$G at $16\times16$ to $35.771$G at $128\times128$, while peak forward memory increases from $19.7$ MiB to $542.8$ MiB. Latency is dominated by launch overhead at resolutions below $64\times64$ and increases to $51.28$ ms per batch at $128\times128$. \autoref{fig:gauge_scalability} is a computational scalability analysis and should not be interpreted as a cross-resolution accuracy experiment.

\section{Additional Method Details}
\label{app:method_details}

\subsection{Physics-Informed Adaptive Allocation}
\label{app:adaptive_allocation_details}

For the latest input slice $\mathbf{v}$, the physical indicators used by the
importance predictor are
\begin{align}
    Q_g(\mathbf{x})
    &=
    \left[
    \frac{1}{C}\sum_{c=1}^{C}
    \left(
    |D_xv^{(c)}(\mathbf{x})|^2+
    |D_yv^{(c)}(\mathbf{x})|^2
    \right)
    \right]^{1/2},
    \nonumber\\
    Q_{\Delta}(\mathbf{x})
    &=
    \frac{1}{C}\sum_{c=1}^{C}
    |D_xD_xv^{(c)}(\mathbf{x})+
    D_yD_yv^{(c)}(\mathbf{x})|,
    \nonumber\\
    Q_E(\mathbf{x})
    &=
    \frac{1}{C}\sum_{c=1}^{C}|v^{(c)}(\mathbf{x})|^2,
    \nonumber\\
    Q_{\omega}(\mathbf{x})
    &=
    |D_xv^{(2)}(\mathbf{x})-D_yv^{(1)}(\mathbf{x})|.
    \label{eq:physical_allocation_indicators}
\end{align}
Here, $\mathbf{x}$ is a spatial location, $\mathbf{v}$ is the latest input
field, $C$ is its number of physical channels, $c$ is the channel index, and
$v^{(c)}(\mathbf{x})$ is the value of channel $c$ at $\mathbf{x}$. The
operators $D_x$ and $D_y$ are finite-difference derivatives along the two
spatial directions. The quantities $Q_g$, $Q_{\Delta}$, $Q_E$, and
$Q_{\omega}$ denote the gradient magnitude, Laplacian magnitude, local energy,
and vorticity-related response, respectively.

The four indicators provide complementary local information. Specifically,
$Q_g$ measures first-order variation, $Q_{\Delta}$ captures second-order local
structure, $Q_E$ measures field strength or activity, and $Q_{\omega}$ provides
a cue related to rotation or shear. For scalar fields, where the
two-component response required by $Q_{\omega}$ is unavailable, the
implementation uses the Laplacian-magnitude channel instead.

As defined in \autoref{eq:importance_prediction}, the indicators are normalized
and concatenated with the learned descriptor $\mathbf{P}$. They serve as
explicit conditioning variables rather than an assumed unique or complete
description of the governing dynamics. The nonlinear predictor may therefore
adjust their relative contributions according to the current input state.

In \autoref{eq:adaptive_displacement}, the mean normalization makes $w_i$ a
relative positive importance weight. The lower bound $I_{\min}$ prevents any
node from being completely suppressed, while $s_I$ and $\delta_{\max}$ control
the importance-modulation range and maximum displacement scale, respectively.
Consequently, the indicators directly condition node relocation rather than
serving only as auxiliary prediction targets.

\subsection{Gauge-Aware Mesh Operator}
\label{app:gauge_operator_details}

To motivate source-to-target transport, let $\mathbf{g}_i$ denote the local
discretization context at node $i$ and let $\mathbf{z}_i$ denote its latent
physical content. We model the encoded feature as
\begin{equation}
    \mathbf{h}_i
    =
    \mathbf{R}(\mathbf{g}_i)\mathbf{z}_i,
    \label{eq:local_frame_model}
\end{equation}
where $\mathbf{h}_i$ is the encoded feature at node $i$,
$\mathbf{g}_i$ collects its local discretization variables,
$\mathbf{z}_i$ represents the underlying latent physical content, and
$\mathbf{R}(\mathbf{g}_i)$ is an invertible representation map associated
with that local context.

Under this representation model, the ideal source-to-target transport is
\begin{equation}
    \mathbf{T}_{i\leftarrow j}^{\star}
    =
    \mathbf{R}(\mathbf{g}_i)
    \mathbf{R}(\mathbf{g}_j)^{-1}.
    \label{eq:ideal_target_transport}
\end{equation}
Here, $\mathbf{T}_{i\leftarrow j}^{\star}$ denotes the ideal transport from
source node $j$ to target node $i$, the superscript $\star$ distinguishes the
ideal transport from its learned approximation, and
$\mathbf{R}(\mathbf{g}_j)^{-1}$ is the inverse representation map of the
source context. The representation maps and local contexts have been defined
in \autoref{eq:local_frame_model}.

Applying this ideal transport to $\mathbf{h}_j$ gives
$\mathbf{R}(\mathbf{g}_i)\mathbf{z}_j$, which expresses the source content
under the target representation context. This relation motivates the learned
transport but is not imposed as an exact equality on the implemented network.

The learned edge descriptor in \autoref{eq:gauge_edge_descriptor} includes
relative adaptive coordinates, source and target mesh Jacobians, and their
state descriptors. It therefore conditions information exchange on both the
geometry produced by adaptive allocation and the current encoded state.

The diagonal component in \autoref{eq:low_rank_transport} supports efficient
channel-wise adjustment, while the gated low-rank component permits
cross-channel correction without predicting an unrestricted
$C_h\times C_h$ matrix for every edge. Separate prediction heads process
$\mathbf{e}_{ij}$ to generate $\mathbf{d}_{ij}^{(\ell)}$,
$\mathbf{U}_{ij}^{(\ell)}$, $\mathbf{V}_{ij}^{(\ell)}$, and
$\gamma_{ij}^{(\ell)}$.

In \autoref{eq:transported_aggregation}, transport is applied before the
weighted sum. The operator therefore exposes how the source--target geometry
conditions each incoming message instead of leaving this dependence implicit
in direct aggregation. The residual update in
\autoref{eq:gauge_residual_update} then combines the transported message with
the existing target representation.

The learned transformation is intended to reduce representation incompatibility
within the considered adaptive-mesh regime. It is not claimed to recover a
unique physical frame, represent an exact analytic coordinate transformation,
or impose exact Gauge equivariance. The corresponding sufficient conditions,
approximation analysis, and deformation-continuity results are presented in
Appendix~\ref{sec:operator_readability_theory}.

\subsection{Mesh-to-Grid Bridge}
\label{sec:mesh_to_grid_bridge}

The gauge-aware operator produces features on adaptive nodes, whereas the output head expects a regular spatial tensor. When the operator and output grids have the same resolution, the bridge assigns distance-based weights to nearby adaptive nodes for each target location $\mathbf{y}^{r}_{q}$:
\begin{equation}
    \kappa_{qi}
    =
    \frac{\exp(-\|\mathbf{x}^{a}_{i}-\mathbf{y}^{r}_{q}\|_2/\tau_b)}
    {\sum_{j\in\mathcal{N}_{b}(q)}\exp(-\|\mathbf{x}^{a}_{j}-\mathbf{y}^{r}_{q}\|_2/\tau_b)}.
\end{equation}
Here, $\kappa_{qi}$ is the reconstruction weight from adaptive node $i$ to target grid point $q$, $\mathbf{y}^{r}_{q}$ is the $q$th regular-grid coordinate, $\mathcal{N}_{b}(q)$ is the bridge neighborhood, and $\tau_b$ is the temperature controlling distance sensitivity.

The reconstructed feature at grid point $q$ is
\begin{equation}
    \mathbf{z}_{q}
    =
    \sum_{i\in\mathcal{N}_{b}(q)}\kappa_{qi}\widehat{\mathbf{h}}_{i},
    \qquad
    \widehat{\mathbf{h}}_{i}=w_i\mathbf{h}_{i}^{(L_g)}.
\end{equation}
Here, $\mathbf{z}_{q}$ is the interpolated grid feature, $\widehat{\mathbf{h}}_{i}$ is the mean-normalized importance-modulated final nodal feature, $w_i$ is defined in the adaptive-mesh module, $L_g$ is the number of gauge-aware layers, and $\mathbf{h}_{i}^{(L_g)}$ is the final feature of adaptive node $i$.

The regular tensor is then obtained by projection and reshaping:
\begin{equation}
    \mathbf{F}_{\mathrm{grid}}
    =
    \operatorname{Reshape}_{H,W}\left(\left[\rho_{b}(\mathbf{z}_{1}),\ldots,\rho_{b}(\mathbf{z}_{M})\right]\right).
\end{equation}
Here, $\rho_b$ is a learnable channel projection, $H$ and $W$ are the target grid height and width, and $M=HW$ is the number of regular-grid locations. In the default two-dimensional configuration, the gauge operator uses stride $2$ and therefore follows the coarse-grid path: its feature map is bilinearly upsampled and then passed through a channel projection. The distance-weighted bridge above is used when the operator and output grids have equal resolution. We consequently use \emph{mesh-to-grid reconstruction} as the general name for these two implementation paths.

\subsection{Spectral and Differential Residual Correction}
\label{sec:pde_aware_correction}

The prediction head combines local convolutional prediction with spectral and physical refinement. The local solver input is
\begin{equation}
    \mathbf{Z}_{\mathrm{sol}}
    =
    \operatorname{Concat}(\mathbf{F}_{\mathrm{grid}},\mathbf{v}),
\end{equation}
where $\mathbf{Z}_{\mathrm{sol}}$ is the solver input and $\mathbf{v}$ denotes the latest input field for temporal prediction or the conditioning field for steady-state prediction. The local solver first produces $\boldsymbol{\delta}_{\mathrm{raw}}=\mathcal{H}_{\mathrm{local}}(\mathbf{Z}_{\mathrm{sol}})$. The spectral branch receives the complete input history, reconstructed grid features, and canonical coordinates:
\begin{align}
    \mathbf{X}_{\mathrm{spec}}
    &=\operatorname{Concat}(\operatorname{Flatten}_{t}(\mathbf{a}),\mathbf{F}_{\mathrm{grid}},\mathbf{X}^{r}),
    \nonumber\\
    \mathbf{Z}_{\mathrm{spec}}
    &=\mathcal{L}_{\mathrm{spec}}(\mathbf{X}_{\mathrm{spec}}),
    \nonumber\\
    \boldsymbol{\delta}_{\mathrm{spec}}
    &=s_{\mathrm{spec}}\mathcal{P}_{\mathrm{spec}}\!\left(\varphi\!\left(\mathcal{N}_{\mathrm{spec}}\!\left(\mathcal{C}_{\mathrm{F}}(\mathbf{Z}_{\mathrm{spec}})+\mathcal{C}_{1\times1}(\mathbf{Z}_{\mathrm{spec}})\right)\right)\right).
\end{align}
Here, $\mathcal{L}_{\mathrm{spec}}$ and $\mathcal{P}_{\mathrm{spec}}$ are the spectral lift and output projection, $\mathcal{C}_{\mathrm{F}}$ is a truncated Fourier convolution retaining a prescribed set of low-order modes, $\mathcal{C}_{1\times1}$ is a pointwise spatial path, $\mathcal{N}_{\mathrm{spec}}$ is group normalization, $\varphi$ is GELU, and $s_{\mathrm{spec}}=0.2\tanh(\beta_{\mathrm{spec}})$ is a learned bounded scale. This module is therefore a global spectral correction, not an explicit high-pass filter.

The intermediate residual $\boldsymbol{\delta}_{1}=\boldsymbol{\delta}_{\mathrm{raw}}+\boldsymbol{\delta}_{\mathrm{spec}}$ is then refined using finite-difference features of its provisional field:
\begin{align}
    \mathbf{u}_{\mathrm{base}}
    &=\mathbf{v}+\boldsymbol{\delta}_{1},
    \nonumber\\
    \mathbf{Z}_{\mathrm{phy}}
    &=\operatorname{Concat}(\mathbf{v},\boldsymbol{\delta}_{1},D_x\mathbf{u}_{\mathrm{base}},D_y\mathbf{u}_{\mathrm{base}},
    \nonumber\\
    &\hspace{31mm}\Delta\mathbf{u}_{\mathrm{base}},E(\mathbf{u}_{\mathrm{base}})),
    \nonumber\\
    \boldsymbol{\delta}
    &=\boldsymbol{\delta}_{1}+s_{\mathrm{phy}}\mathcal{H}_{\mathrm{phy}}(\mathbf{Z}_{\mathrm{phy}}).
\end{align}
Here, $\mathcal{H}_{\mathrm{phy}}$ is the convolutional differential-correction network, $s_{\mathrm{phy}}=0.2\tanh(\beta_{\mathrm{phy}})$ is its learned bounded scale, $\Delta$ is the discrete Laplacian, and $E$ is the channel-averaged squared amplitude. For direct-output tasks in which the conditioning and output channel counts differ, $\mathbf{u}_{\mathrm{base}}$ is set to $\boldsymbol{\delta}_{1}$.

The final prediction is
\begin{equation}
    \widehat{\mathbf{u}}
    =
    \begin{cases}
    \mathbf{u}_{t}+\boldsymbol{\delta}, & \text{for temporal residual prediction},\\
    \boldsymbol{\delta}, & \text{for steady-state prediction or direct output}.
    \end{cases}
\end{equation}
In this equation, $\mathbf{u}_{t}$ is the latest observed state and $\widehat{\mathbf{u}}$ is the predicted future or steady-state field. All modules are trained end to end, with mesh-quality and transport-consistency terms used as auxiliary regularization during training.

\section{Theoretical Analysis of Operator Readability}
\label{sec:operator_readability_theory}

We formalize \emph{operator readability} relative to a declared family of nondegenerate adaptive discretizations. It has two requirements. First, allocation readability requires the node displacement to be an explicit function of specified input and geometry variables, so that their influence can be inspected or intervened on. Second, interaction readability requires a source message to have the same physical meaning after an admissible change of its local representation convention. The first requirement concerns where computation is allocated; the analysis below focuses on the second requirement, which concerns how the resulting representations are exchanged.

Let $\mathbf{g}_i$ collect the local discretization variables at node $i$, including local coordinates, mesh Jacobians, and state descriptors. The representation model in \autoref{eq:local_frame_model} adopts the standard gauge-equivariant viewpoint that the same underlying content may have different coordinate representations under changes of local frame \citep{cohen2019gauge,dehaan2021gauge}. Here, we adapt this viewpoint to input-dependent discretizations without assuming that the implemented GA-AMNO satisfies exact gauge equivariance. Specifically, $\mathbf{z}_i\in\mathbb{R}^{C_h}$ is fixed latent physical content and $\mathbf{R}(\mathbf{g}_i)\in\mathfrak{G}\subseteq\mathrm{GL}(C_h)$ is invertible. The admissible group $\mathfrak{G}$ contains only representation changes induced by the considered nondegenerate mesh family; it is not identified with every element of $\mathrm{GL}(C_h)$. We assume that $\mathbf{R}$ and $\mathbf{R}^{-1}$ are bounded on this family. Unless the Frobenius norm is explicitly indicated, matrix norms below are induced $\ell_2$ norms. This representation model is our formalization for the stated adaptive-discretization setting; it is sufficient for the analysis and does not claim that the frame is unique or directly observable.

\noindent\textbf{Definition 1 (Discretization-induced representation ambiguity).} For fixed latent physical content $\mathbf{z}_i$, define its admissible representation set as
\begin{equation}
\mathcal{O}(\mathbf{z}_i)
=
\left\{
\mathbf{R}(\mathbf{g})\mathbf{z}_i:
\mathbf{g}\in\mathfrak{D}
\right\},
\label{eq:representation_orbit}
\end{equation}
where $\mathfrak{D}$ is the declared family of nondegenerate local discretization contexts. Discretization-induced representation ambiguity occurs when $\mathcal{O}(\mathbf{z}_i)$ contains more than one element, so that the same physical content admits different encoded features under different admissible local discretizations. For two contexts $\mathbf{g}_i,\mathbf{g}'_i\in\mathfrak{D}$, the corresponding representations satisfy
\begin{equation}
\mathbf{h}'_i
=
\mathbf{S}_i\mathbf{h}_i,
\qquad
\mathbf{S}_i
=
\mathbf{R}(\mathbf{g}'_i)
\mathbf{R}(\mathbf{g}_i)^{-1}.
\label{eq:ambiguity_reparameterization}
\end{equation}
We refer to this local nonuniqueness of feature coordinates, with $\mathbf{z}_i$ unchanged, as a hidden gauge ambiguity. This is a modeling description of admissible representation changes, not a claim that the implemented network is exactly gauge equivariant.

\noindent\textbf{Definition 2 (Interaction readability).} A directed interaction $j\rightarrow i$ satisfies interaction readability if its message depends on the source physical content $\mathbf{z}_j$ and the target context $\mathbf{g}_i$, but not on the chosen admissible source-frame convention used to encode $\mathbf{z}_j$. Equivalently, changing local representation conventions must transform the aggregated message according to the target action $\mathbf{S}_i$, rather than according to the independently varying source actions. An adaptive mesh is operator-readable relative to the declared discretization family when this interaction property is combined with allocation readability.

\noindent\textbf{Theorem 1 (Limitation of shared direct aggregation).} Consider the uncorrected linear message
\begin{equation}
\overline{\mathbf{m}}_i
=
\sum_{j\in\mathcal{N}(i)}
\alpha_{ij}\mathbf{W}\mathbf{h}_j,
\label{eq:direct_aggregation}
\end{equation}
where $\mathbf{W}$ is shared across edges. Consider the subset of local reparameterizations $\mathbf{h}'_j=\mathbf{S}_j\mathbf{h}_j$ that leaves the scalar weights unchanged. If $\overline{\mathbf{m}}'_i=\mathbf{S}_i\overline{\mathbf{m}}_i$ must hold for all source features, then every edge with $\alpha_{ij}\neq0$ must satisfy
\begin{equation}
\mathbf{W}\mathbf{S}_j=\mathbf{S}_i\mathbf{W}.
\label{eq:shared_intertwining_condition}
\end{equation}
Consequently, a single shared map cannot guarantee operator-readable aggregation for independently varying local frames.

\noindent\textit{Proof:} The required equality is
\begin{equation}
\sum_j\alpha_{ij}\mathbf{W}\mathbf{S}_j\mathbf{h}_j
=
\sum_j\alpha_{ij}\mathbf{S}_i\mathbf{W}\mathbf{h}_j.
\end{equation}
Because the source features can vary independently, each nonzero edge must satisfy \autoref{eq:shared_intertwining_condition}. For $\mathbf{W}=\mathbf{I}$, this reduces to $\mathbf{S}_j=\mathbf{S}_i$; for invertible $\mathbf{W}$, it imposes the same fixed conjugacy relation on every neighbor. Neither condition accommodates independently changing local discretization frames in general.

Theorem 1 concerns shared uncorrected aggregation, not every irregular-mesh operator. A geometry-conditioned edge map may avoid this limitation; such a map already acts as a form of representation transport.

\noindent\textbf{Theorem 2 (Sufficient condition for operator-readable transport).} Let $\alpha_{ij}\geq0$ and $\sum_j\alpha_{ij}=1$. If $\mathbf{T}_{i\leftarrow j}=\mathbf{T}_{i\leftarrow j}^{\star}$, with the ideal transport defined in \autoref{eq:ideal_target_transport}, then the transported aggregation satisfies
\begin{equation}
\mathbf{m}_i
=
\sum_j\alpha_{ij}
\mathbf{T}_{i\leftarrow j}\mathbf{h}_j
=
\mathbf{R}(\mathbf{g}_i)
\sum_j\alpha_{ij}\mathbf{z}_j.
\label{eq:operator_readable_target_aggregate}
\end{equation}
Moreover, under a local coordinate change $\mathbf{h}'_i=\mathbf{S}_i\mathbf{h}_i$, if $\alpha'_{ij}=\alpha_{ij}$ and
\begin{equation}
\mathbf{T}'_{i\leftarrow j}
=
\mathbf{S}_i\mathbf{T}_{i\leftarrow j}\mathbf{S}_j^{-1},
\label{eq:connection_covariance_theory}
\end{equation}
then $\mathbf{m}'_i=\mathbf{S}_i\mathbf{m}_i$.

\noindent\textit{Proof:} Substituting \autoref{eq:local_frame_model} and \autoref{eq:ideal_target_transport} cancels the source-frame factors before summation and gives \autoref{eq:operator_readable_target_aggregate}. Under the coordinate change,
\begin{equation}
\mathbf{m}'_i
=
\sum_j\alpha_{ij}
\mathbf{S}_i\mathbf{T}_{i\leftarrow j}
\mathbf{S}_j^{-1}\mathbf{S}_j\mathbf{h}_j
=
\mathbf{S}_i\mathbf{m}_i.
\end{equation}
Thus, all incoming physical contents are represented in one target frame, and changing local coordinates changes only the coordinates of the target message.

Theorem 2 is an exact sufficient condition. The learned diagonal-plus-low-rank map is not asserted to satisfy it identically. To cover the learned case, define the attention defect and connection defect as
\begin{align}
\delta_{\alpha,i}
&=
\sum_j|\alpha'_{ij}-\alpha_{ij}|,
\nonumber\\
\varepsilon_{ij}
&=
\left\|
\mathbf{T}'_{i\leftarrow j}\mathbf{S}_j
-\mathbf{S}_i\mathbf{T}_{i\leftarrow j}
\right\|_2.
\label{eq:learned_consistency_defects}
\end{align}

\noindent\textbf{Theorem 3 (Approximate interaction-consistency bound).} Assume $\|\mathbf{h}_j\|_2\leq H$ and $\|\mathbf{T}'_{i\leftarrow j}\mathbf{S}_j\|_2\leq B$. Then
\begin{equation}
\left\|
\mathbf{m}'_i-\mathbf{S}_i\mathbf{m}_i
\right\|_2
\leq
BH\delta_{\alpha,i}
+
H\sum_j\alpha_{ij}\varepsilon_{ij}.
\label{eq:approximate_gauge_bound}
\end{equation}

\noindent\textit{Proof:} Add and subtract $\sum_j\alpha_{ij}\mathbf{T}'_{i\leftarrow j}\mathbf{S}_j\mathbf{h}_j$ to obtain
\begin{align}
\mathbf{m}'_i-\mathbf{S}_i\mathbf{m}_i
&=\sum_j(\alpha'_{ij}-\alpha_{ij})
\mathbf{T}'_{i\leftarrow j}\mathbf{S}_j\mathbf{h}_j
\nonumber\\
&\quad+\sum_j\alpha_{ij}
(\mathbf{T}'_{i\leftarrow j}\mathbf{S}_j
-\mathbf{S}_i\mathbf{T}_{i\leftarrow j})\mathbf{h}_j.
\end{align}
The triangle inequality and submultiplicativity bound the first sum by $BH\delta_{\alpha,i}$ and the second by $H\sum_j\alpha_{ij}\varepsilon_{ij}$.

The bound separates two failure modes of a learned layer: the scalar neighbor weighting may change, and the feature transport may deviate from connection covariance. It therefore applies even when attention is produced from learned state descriptors rather than assumed to be exactly gauge invariant.

\noindent\textbf{Proposition 1 (Approximation capacity of diagonal-plus-low-rank transport).}
Let $\mathbf{T}^{\star}\in\mathbb{R}^{C_h\times C_h}$ denote an ideal transport matrix for one edge, and let $\mathbf{D}$ be any admissible diagonal matrix. Define $\mathbf{E}=\mathbf{T}^{\star}-\mathbf{D}$, with singular values $\sigma_1(\mathbf{E})\geq\cdots\geq\sigma_{C_h}(\mathbf{E})$. Then the best rank-$r$ correction satisfies
\begin{equation}
\min_{\operatorname{rank}(\mathbf{L})\leq r}
\left\|\mathbf{T}^{\star}-(\mathbf{D}+\mathbf{L})\right\|_F^2
=
\sum_{k>r}\sigma_k^2(\mathbf{E}).
\label{eq:low_rank_transport_capacity}
\end{equation}
Consequently, the parameterization $\mathbf{D}+\mathbf{U}\mathbf{V}^{\top}$ can represent the optimal rank-$r$ approximation of the non-diagonal remainder for the selected $\mathbf{D}$.

\noindent\textit{Proof:} For fixed $\mathbf{D}$, the problem is equivalent to finding the best rank-$r$ approximation to $\mathbf{E}$. \autoref{eq:low_rank_transport_capacity} follows from the Eckart--Young--Mirsky theorem \citep{eckart1936approximation,mirsky1960symmetric}, and a truncated singular value decomposition supplies factors $\mathbf{U}$ and $\mathbf{V}$. The edgewise scalar gate and normalization used in the implementation can be absorbed into these unconstrained factors whenever the gate is nonzero.

Proposition 1 is an expressivity statement, not an optimization guarantee. It explains why the diagonal term can model channelwise rescaling while a small rank captures dominant cross-channel frame mixing. The residual tail in \autoref{eq:low_rank_transport_capacity} contributes to the connection defect $\varepsilon_{ij}$ in Theorem 3; the learned edge network is not assumed to attain the best approximation.

\noindent\textbf{Proposition 2 (Topology-preserving deformation stability).} Consider two discretizations $\mathcal{X}$ and $\mathcal{X}'$ of the same physical input, with the same node identities and neighborhood topology, and let $\delta_{\mathcal{X}}=\max_i\|\mathbf{x}'_i-\mathbf{x}_i\|_2$. If all geometry-dependent learned features, normalized edge weights, and induced transport matrices are locally Lipschitz functions of the coordinates on a compact family of nondegenerate meshes, then there is a finite $L_m$ such that
\begin{equation}
\|\mathbf{m}'_i-\mathbf{m}_i\|_2
\leq
L_m\delta_{\mathcal{X}}.
\label{eq:message_mesh_continuity}
\end{equation}

\noindent\textit{Justification:} For the same physical sample, decompose the message difference into feature, weight, and transport changes and apply their local Lipschitz bounds. This gives \autoref{eq:message_mesh_continuity}. The statement applies only while connectivity is fixed and cells remain nondegenerate; it does not cover folding or topology changes.

\noindent\textbf{Corollary 1 (Propagation to the operator output).}
Let $\Phi_{\ell}$ and $\Phi'_{\ell}$ denote the $\ell$th complete feature-update maps under two admissible descriptions of the same physical input, and let $\mathbf{S}_{\ell}$ denote the corresponding action on layer-$\ell$ features. Assume
\begin{equation}
\left\|
\Phi'_{\ell}(\mathbf{S}_{\ell-1}\mathbf{h})
-\mathbf{S}_{\ell}\Phi_{\ell}(\mathbf{h})
\right\|_2
\leq b_{\ell}
\label{eq:layer_intertwining_defect}
\end{equation}
on the considered compact domain, where $b_{\ell}$ can be bounded using Theorem 3 together with the deformation contribution in Proposition 2. Suppose $\Phi'_{q}$ is $K_q$-Lipschitz. For the complete reconstruction and prediction maps $\mathcal{D}$ and $\mathcal{D}'$, assume that $\mathcal{D}'$ is $K_{\mathrm{out}}$-Lipschitz and
\begin{equation}
\left\|
\mathcal{D}'(\mathbf{S}_{L_g}\mathbf{h})
-\mathbf{S}_{\mathrm{out}}\mathcal{D}(\mathbf{h})
\right\|_2
\leq b_{\mathrm{out}}.
\label{eq:decoder_intertwining_defect}
\end{equation}
Then the final predictions satisfy
\begin{equation}
\left\|\widehat{\mathbf{u}}'-\mathbf{S}_{\mathrm{out}}\widehat{\mathbf{u}}\right\|_2
\leq
K_{\mathrm{out}}
\sum_{\ell=1}^{L_g}
\left(\prod_{q=\ell+1}^{L_g}K_q\right)b_{\ell}
+b_{\mathrm{out}}.
\label{eq:end_to_end_consistency_bound}
\end{equation}
Here, $\mathbf{S}_{\mathrm{out}}$ is the prescribed output action; it is the identity after reconstruction to the same canonical grid. An empty product in \autoref{eq:end_to_end_consistency_bound} is defined as one.

\noindent\textit{Proof:} Apply \autoref{eq:layer_intertwining_defect} at the first layer and propagate its defect through each later Lipschitz map. Repeating this argument for every layer and using the triangle inequality gives the weighted sum in \autoref{eq:end_to_end_consistency_bound}. Applying \autoref{eq:decoder_intertwining_defect} contributes the final term $b_{\mathrm{out}}$.

Corollary 1 connects local transport quality to final operator stability, but it also exposes the remaining requirements: small edgewise defects alone are insufficient if later mappings have large Lipschitz constants or are incompatible with the canonical output representation. In the implemented architecture, the transport-continuity and loop terms are auxiliary regularizers rather than exact constraints. Moreover, for the low-rank mode the loop term regularizes the diagonal channel-scaling component only; no exact holonomy claim is made for the full diagonal-plus-low-rank transport.

This analysis is performed on the induced matrix $\mathbf{T}_{i\leftarrow j}$, not on the individual diagonal and low-rank factors. It establishes the limitation of shared direct aggregation, an exact sufficient condition for operator-readable transport, an approximation-capacity result for the implemented parameterization, and local-to-output bounds for the learned approximate case. It does not claim unique recovery of a physical frame or exact invariance of the implemented network. The intervention and deformation studies in \autoref{sec:experimental_results} examine whether the learned connection is functionally used and remains stable, while the message--target discrepancy reported there is treated as a compatibility diagnostic rather than a direct measurement of the latent covariance defect in \autoref{eq:learned_consistency_defects}.

\section{Expanded Results}
\subsection{Targeted Component Ablations}
\label{sec:extended_component_ablation}

\begin{figure*}[!t]
\centering
\includegraphics[width=0.92\textwidth]{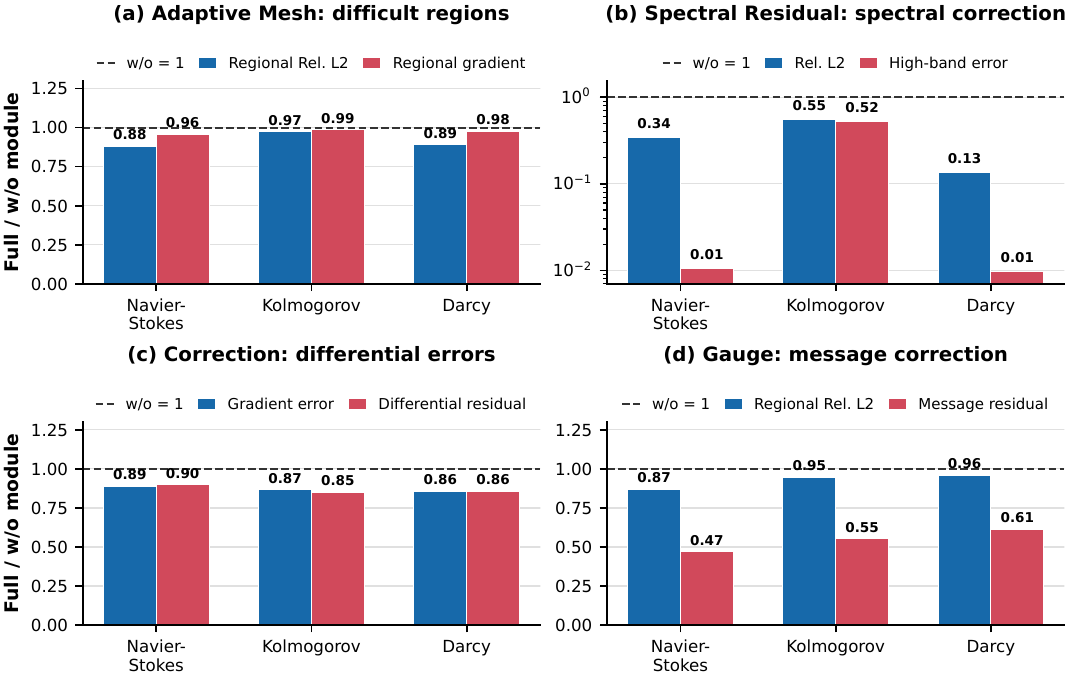}
\caption{Targeted component ablations using frozen checkpoints. Each bar is the Full GA-AMNO metric divided by the corresponding w/o-module metric; lower is better and the dashed line denotes the ablated reference. Adaptive Mesh is evaluated in high-demand regions, Spectral Residual through field and high-band spectral errors, Differential Correction through gradient and differential errors, and Low-rank Gauge Transport through regional prediction and residual message--target discrepancy. Panel (b) uses a logarithmic vertical axis.}
\label{fig:targeted_module_ablation}
\end{figure*}

We remove one component at a time while retaining the remaining training protocol and evaluate each ablation through the failure mode that the removed component is intended to address. All comparisons use frozen checkpoints. To make quantities with different scales visually comparable, \autoref{fig:targeted_module_ablation} reports the ratio between the Full GA-AMNO error and the corresponding w/o-module error. The dashed line at one is the ablated reference; a value below one means that the complete model is better.

\textbf{1) Adaptive Mesh: difficult-region allocation.} We use the target-free high-demand masks defined in \autoref{sec:experimental_setting} and measure regional relative $\ell_2$ and gradient errors. This test asks whether adaptive allocation improves the regions that exhibit strong input variation and nonuniform geometric demand, rather than whether it uniformly changes the whole field.

\textbf{2) Spectral Residual: spectral correction.} We jointly measure full-field relative $\ell_2$ error and high-band spectral-energy error. The high band contains Fourier coefficients whose normalized radial frequency is at least $0.6$ of the maximum represented radius. Although the implemented Fourier convolution retains truncated low-order modes rather than applying an explicit high-pass mask, this diagnostic tests whether the complete spectral-plus-pointwise correction empirically improves frequency content that is difficult for the mesh-domain prediction alone.

\textbf{3) Differential Correction: local-structure repair.} We measure gradient error and the dataset-specific differential residual used in the evaluation pipeline. For Darcy, the latter is the Laplacian discrepancy defined in \autoref{sec:experimental_setting}, not a claim of evaluating the complete coefficient-weighted equation residual. These metrics match the derivative features supplied to the correction head.

\textbf{4) Low-rank Gauge Transport: geometry-conditioned message correction.} We measure high-demand regional relative $\ell_2$ error together with residual message--target discrepancy after transport. For the compatibility test, Full GA-AMNO and w/o Low-rank Gauge Transport receive the same GA-AMNO adaptive mesh and identical smooth perturbations, isolating the low-rank correction from mesh allocation.

All bars in \autoref{fig:targeted_module_ablation} are below one. Adaptive allocation reduces difficult-region errors by approximately $1\%$--$12\%$. The spectral correction gives the largest targeted effect, reducing relative $\ell_2$ and high-band errors by $45\%$--$99\%$; this is an empirical ablation result rather than a claim that the branch is structurally restricted to high frequencies. Differential correction reduces its two local-structure diagnostics by $10\%$--$15\%$. Low-rank Gauge Transport reduces high-demand prediction error by $4\%$--$13\%$ and residual message--target discrepancy by $39\%$--$53\%$. The significance of this experiment is the observed module--failure-mode correspondence: removing a component primarily degrades the quantity that the component was designed to control. This correspondence shows that the architecture is not merely an undifferentiated stack of accuracy modules and links Physics-Informed Adaptive Allocation and Low-rank Gauge Transport to their intended computational roles. It does not imply that every component lowers every global metric on every dataset.

\subsection{Smooth Adaptive-Mesh Perturbations}
\label{sec:smooth_mesh_perturbations}

We next test whether the learned message-correction mechanism remains stable after controlled changes to the internally learned mesh. Starting from mesh $G$, we construct
\begin{equation}
G_{k,\ell}=G+\ell\eta d_k(G), \qquad \ell\in\{0,0.25,0.50,0.75,1.00\},
\label{eq:smooth_mesh_perturbation_revised}
\end{equation}
where $k$ indexes one of six smooth boundary-tapered transformations, $\ell$ is the perturbation strength, $d_k$ is the displacement field, and $\eta$ equals $12\%$ of the mean edge length. The transformations comprise smooth horizontal translation, diagonal translation, rotation, sine-modulated rotation, cosine-modulated rotation, and a sine--cosine twist. Connectivity is fixed and every evaluated mesh has a positive cell Jacobian.

Full GA-AMNO is compared with the trained adaptive-mesh ablation without Low-rank Gauge Transport. Both models receive the same GA-AMNO adaptive mesh and the same perturbation, isolating the effect of Low-rank Gauge Transport from resource allocation.

\begin{table}[!t]
\centering
\caption{Mean message-compatibility gain over six smooth transformations and four nonzero strengths.}
\label{tab:smooth_mesh_perturbation_revised}
\scriptsize
\setlength{\tabcolsep}{4pt}
\renewcommand{\arraystretch}{1.10}
\begin{tabular}{lccc}
\hline
Dataset & GA-AMNO (\%) & Adaptive Mesh (\%) & Difference \\
\hline
Navier-Stokes & \textbf{83.15} & 64.13 & +19.02 \\
Kolmogorov & \textbf{48.05} & 6.24 & +41.81 \\
Darcy & \textbf{39.52} & 1.35 & +38.17 \\
\hline
\end{tabular}
\end{table}

As shown in \autoref{tab:smooth_mesh_perturbation_revised}, GA-AMNO has higher message-compatibility gain in all $72$ dataset--transformation--strength combinations. It also has lower prediction error than the adaptive-mesh ablation in all $24$ perturbation conditions on Navier-Stokes and Kolmogorov, but not on Darcy. Moreover, it is not uniformly less sensitive according to a target-free output-change metric. Unlike the previous experiment, which analyzes mismatch already present in the learned mesh, this controlled perturbation test changes the geometry while holding the physical sample and mesh allocation shared between the two models. It therefore verifies that the advantage of low-rank transport persists under multiple smooth, topology-preserving changes of the discretization and is not only a correlation observed on the original mesh. The result supports deformation-robust interaction correction, rather than universal invariance to arbitrary geometry changes.

\subsection{High-Mismatch Message Compatibility}
\label{sec:high_mismatch_compatibility}

To complement the mismatch-stratified results in
\autoref{fig:feature_alignment_revised}, we report the absolute compatibility
measurements for the high-geometric-mismatch group in
\autoref{tab:feature_alignment_revised}. Here,
$E_{\mathrm{src}}$ measures the discrepancy between the untransported source
feature and the target representation, while $E_{\mathrm{msg}}$ measures the
discrepancy after source-to-target transport. The compatibility gain is
computed as
\[
\mathrm{Gain}
=
\frac{E_{\mathrm{src}}-E_{\mathrm{msg}}}
     {E_{\mathrm{src}}}
\times 100\%.
\]
Transport consistently lowers message--target discrepancy, with gains ranging
from $48.32\%$ to $81.93\%$. These results provide the absolute values underlying
the stratified comparison in the main text.

\begin{table}[!t]
\centering
\caption{Message--target compatibility on high-geometric-mismatch edges.
Lower $E_{\mathrm{src}}$ and $E_{\mathrm{msg}}$ indicate smaller discrepancy,
whereas a larger positive Gain indicates a stronger improvement produced by
low-rank Gauge transport.}
\label{tab:feature_alignment_revised}
\scriptsize
\setlength{\tabcolsep}{5pt}
\renewcommand{\arraystretch}{1.10}
\begin{tabular}{lccc}
\hline
Dataset & $E_{\mathrm{src}}$ & $E_{\mathrm{msg}}$ & Gain (\%) \\
\hline
Navier--Stokes & 0.052823 & 0.009543 & 81.93 \\
Kolmogorov     & 0.127806 & 0.066045 & 48.32 \\
Darcy          & 0.287031 & 0.123934 & 56.82 \\
\hline
\end{tabular}
\end{table}

\subsection{Mesh Quality}
\label{sec:mesh_quality_revised}

Finally, \autoref{tab:mesh_quality_revised} verifies that the learned allocation is realized on valid geometries. KS is omitted because it is one-dimensional. All evaluated two-dimensional meshes have zero inversion rate, minimum cell angles above $84^{\circ}$, and aspect ratios close to one. This experiment rules out a degenerate explanation in which prediction gains are obtained through folded cells, invalid neighborhoods, or extreme element anisotropy. These quantities are geometric feasibility controls rather than evidence of operator readability by themselves; the counterfactual and transport experiments above establish whether the valid adaptive geometry and its interactions are functionally meaningful.

\begin{table}[!t]
\centering
\caption{Geometric quality of the learned adaptive mesh.}
\label{tab:mesh_quality_revised}
\scriptsize
\setlength{\tabcolsep}{3.5pt}
\renewcommand{\arraystretch}{1.12}
\begin{tabular}{lccc}
\hline
Dataset & Inversion Rate & Min. Angle & Aspect Ratio \\
\hline
Navier-Stokes & 0.0000 & $89.019^{\circ}$ & 1.0036 \\
Rayleigh-B\'enard & 0.0000 & $84.308^{\circ}$ & 1.0360 \\
Kolmogorov & 0.0000 & $89.716^{\circ}$ & 1.0038 \\
Darcy & 0.0000 & $89.669^{\circ}$ & 1.0014 \\
\hline
\end{tabular}
\end{table}

Taken together, the experiments support a layered conclusion rather than relying on prediction error alone. The main comparison establishes operator utility; component ablation assigns improvements to intended failure modes; importance counterfactuals test allocation readability; edge intervention and mismatch stratification test whether representation correction is both consequential and geometrically targeted; controlled deformation evaluates whether that mechanism persists when the discretization changes; and mesh diagnostics exclude invalid geometry as an explanation. This chain of evidence supports the central claim that operator readability increases when both node allocation and cross-node interaction can be inspected and functionally tested. It does not establish exact gauge covariance, unique recovery of latent physical frames, or invariance to arbitrary external meshes.

\subsection{Cross-Resolution Generalization}
\autoref{tab:cross_resolution} evaluates the proposed model under resolution changes. The model is trained at the canonical $64 \times 64$ resolution and directly evaluated at resolutions from $16$ to $128$. Only relative $\ell_2$ error is reported to focus on the overall reconstruction robustness under discretization changes.

The lowest error is consistently obtained at the native resolution. Nevertheless, the model remains accurate at nearby resolutions. For example, the Navier-Stokes error is $0.013443$ at resolution $48$ and $0.009215$ at resolution $96$, compared with $0.003219$ at the native resolution. Similar trends are observed for Rayleigh-B\'enard convection, Kuramoto--Sivashinsky, and Kolmogorov flow.

The error increases more noticeably at extremely coarse resolutions, especially for Kolmogorov flow and Darcy flow. This behavior is expected because $16 \times 16$ sampling cannot represent localized vortices, sharp gradients, or heterogeneous permeability patterns adequately. The results indicate that the learned gauge-aware representation is stable under moderate resolution shifts, while its accuracy remains fundamentally tied to the information available in the input discretization.

\begin{figure*}[!t]
\centering
\includegraphics[width=0.98\textwidth]{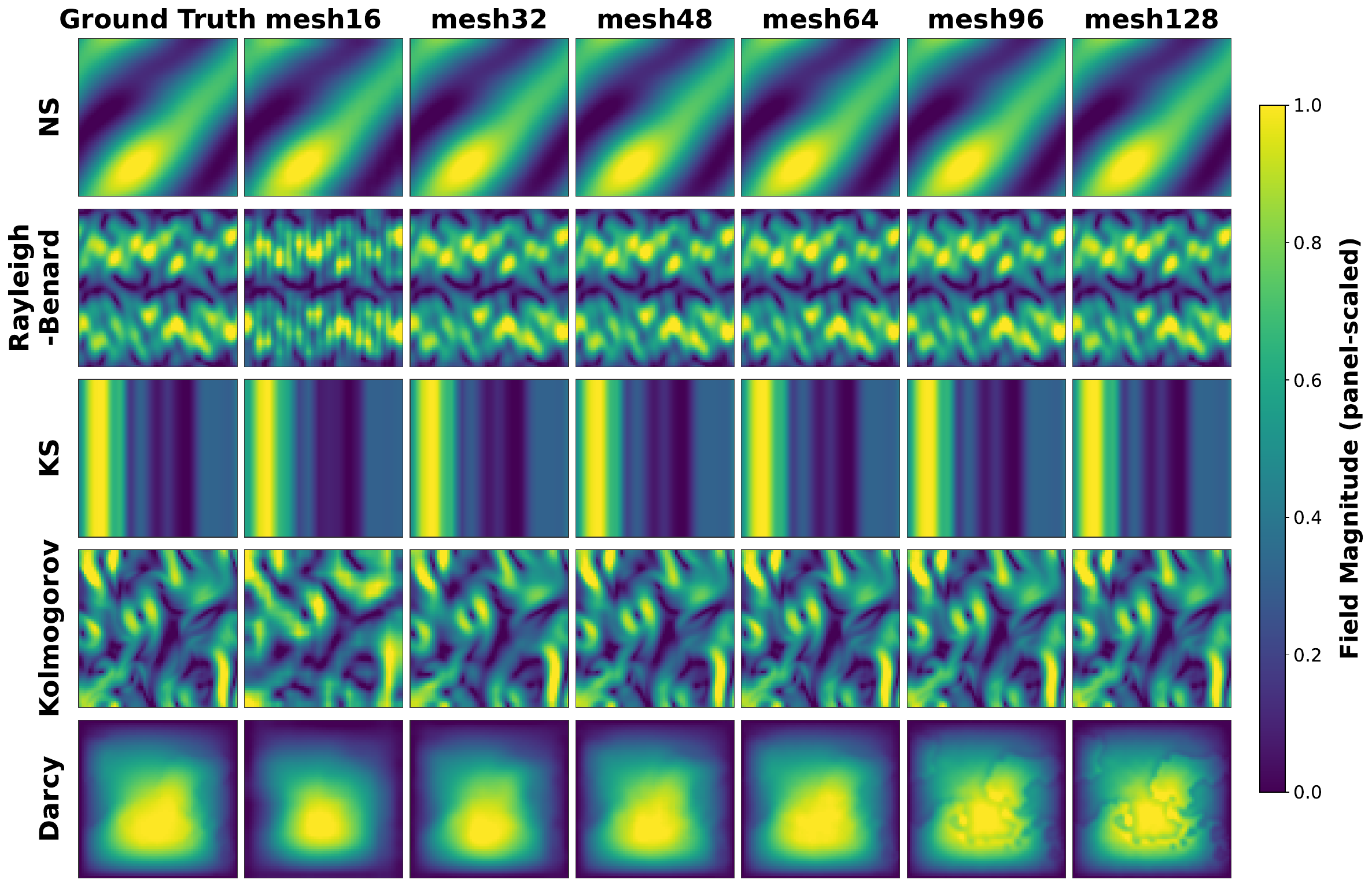}
\caption{GA-AMNO predictions under cross-resolution evaluation. Columns show the ground truth and predictions at mesh resolutions $16$, $32$, $48$, $64$, $96$, and $128$; rows correspond to the five PDE benchmarks. The native training resolution is $64$, and colors are independently scaled within each panel.}
\label{fig:qualitative_cross_resolution}
\end{figure*}

To complement the quantitative results in \autoref{tab:cross_resolution}, \autoref{fig:qualitative_cross_resolution} shows representative predictions across the evaluated resolutions. The visualizations confirm the same resolution-dependent pattern: severe downsampling removes localized and high-frequency structures, whereas resolutions close to or finer than the training resolution preserve the dominant field morphology. Because each panel is scaled independently, the figure is intended to illustrate structural preservation rather than absolute-amplitude agreement or resolution invariance.

\begin{table}[!t]
\centering
\caption{Cross-resolution generalization of GA-AMNO\@. The model is trained at $64 \times 64$. Relative $\ell_2$ error is reported; lower is better.}
\label{tab:cross_resolution}
\scriptsize
\setlength{\tabcolsep}{2.5pt}
\renewcommand{\arraystretch}{1.10}
\begin{tabular}{lcccccc}
\hline
Dataset & $16$ & $32$ & $48$ & $64$ & $96$ & $128$ \\
\hline
Navier-Stokes
& 0.175885 & 0.041609 & 0.013443 & \textbf{0.003219} & 0.009215 & 0.012195 \\
Rayleigh-B\'enard
& 0.202487 & 0.068442 & 0.033256 & \textbf{0.000680} & 0.019584 & 0.020235 \\
KS
& 0.172811 & 0.077675 & 0.045117 & \textbf{0.000387} & 0.013415 & 0.007378 \\
Kolmogorov
& 0.431518 & 0.169014 & 0.064861 & \textbf{0.018796} & 0.058290 & 0.076400 \\
Darcy
& 0.351837 & 0.132896 & 0.061258 & \textbf{0.027743} & 0.083502 & 0.113239 \\
\hline
\end{tabular}
\end{table}

\subsection{Rollout Stability}

\autoref{tab:rollout} reports autoregressive rollout results of GA-AMNO\@. Darcy flow is not included because it is a static PDE operator-learning problem. We report prediction errors at one, five, and ten rollout steps.

For Navier-Stokes, the relative $\ell_2$ error increases from $0.003213$ at one step to $0.030052$ after ten steps. Rayleigh-B\'enard convection remains particularly stable, with a ten-step error of only $0.012066$. Kuramoto--Sivashinsky also maintains a low ten-step error of $0.004134$, despite its chaotic temporal evolution. Kolmogorov flow has the largest long-horizon error, increasing from $0.018522$ to $0.253117$, which reflects the accumulation of phase and vortex-location errors in forced turbulent dynamics.

These results suggest that the proposed model preserves stable temporal evolution for moderate rollout horizons. The differential residual correction helps suppress local high-frequency drift, while Physics-Informed Adaptive Allocation and Low-rank Gauge Transport reduce the propagation of representation mismatch across time. The remaining degradation on Kolmogorov flow highlights that strongly forced turbulence remains challenging for purely autoregressive neural operators.

\begin{figure*}[!t]
\centering
\includegraphics[width=0.88\textwidth]{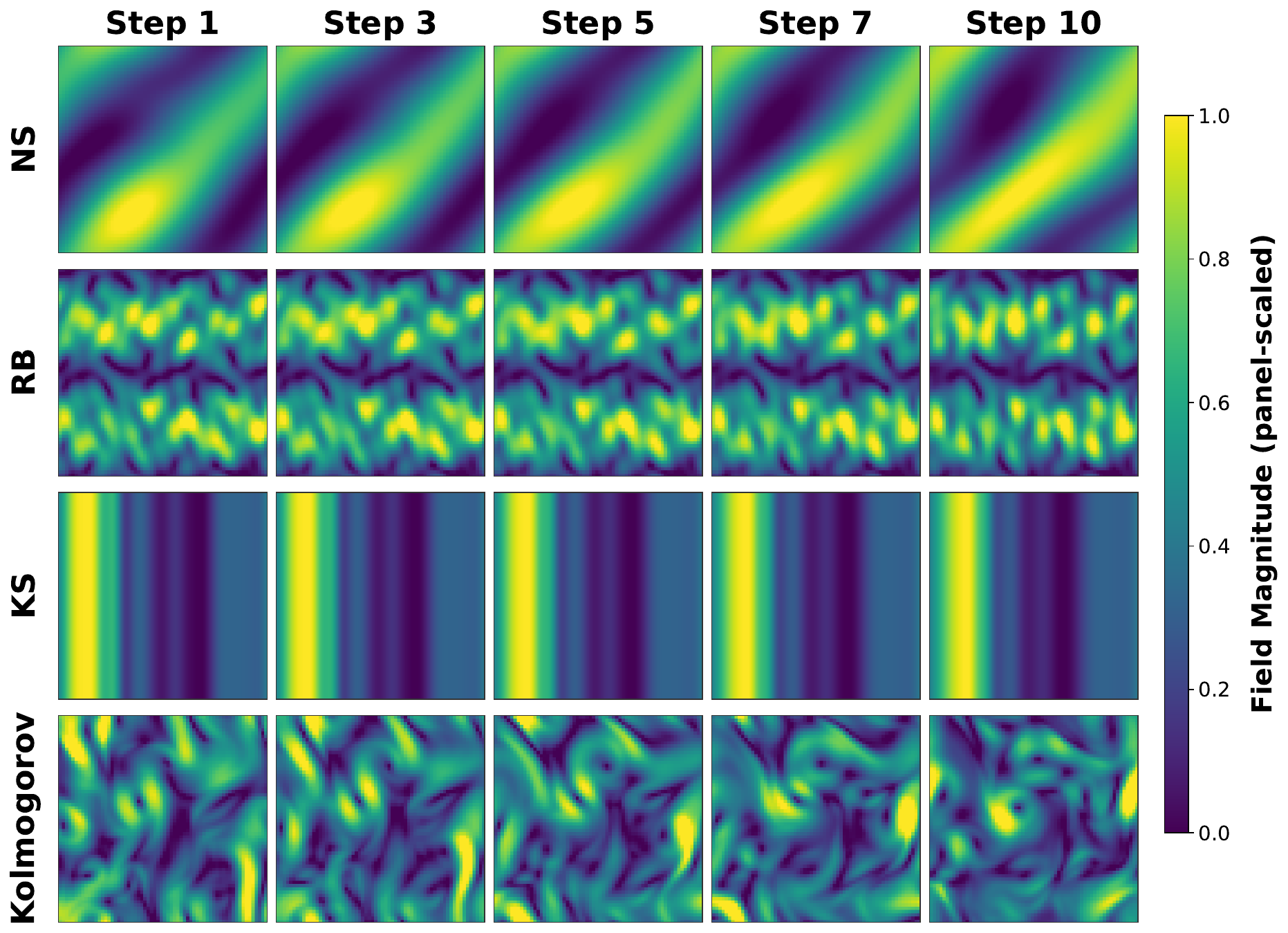}
\caption{Selected GA-AMNO autoregressive predictions at rollout steps $1$, $3$, $5$, $7$, and $10$ for Navier-Stokes, Rayleigh-B\'enard, KS, and Kolmogorov. Colors are independently scaled within each panel, and the one-dimensional KS profile is repeated vertically for visualization.}
\label{fig:qualitative_rollout}
\end{figure*}

The qualitative rollouts in \autoref{fig:qualitative_rollout} provide a visual counterpart to the error growth reported in \autoref{tab:rollout}. Navier-Stokes, Rayleigh-B\'enard, and KS retain their dominant structures over the displayed horizon, whereas Kolmogorov flow exhibits increasingly visible phase and vortex-location deviations. This behavior is consistent with its faster quantitative error accumulation. Darcy is excluded because it defines a steady coefficient-to-solution mapping rather than a temporal rollout task.

\begin{table}[!t]
\centering
\caption{Autoregressive rollout results of GA-AMNO\@. Lower is better. `Spec.' denotes the overall spectral-energy error $\mathrm{SpectrumError}$.}
\label{tab:rollout}
\scriptsize
\setlength{\tabcolsep}{3.5pt}
\renewcommand{\arraystretch}{1.10}
\begin{tabular}{lcccc}
\hline
Dataset & Step & Rel. L2 & RMSE & Spec. \\
\hline
Navier-Stokes & 1  & 0.003213 & 0.002604 & 0.001566 \\
Navier-Stokes & 5  & 0.016251 & 0.014324 & 0.011963 \\
Navier-Stokes & 10 & 0.030052 & 0.027607 & 0.024469 \\
\hline
Rayleigh-B\'enard & 1  & 0.000692 & 0.000085 & 0.000195 \\
Rayleigh-B\'enard & 5  & 0.004747 & 0.000612 & 0.001478 \\
Rayleigh-B\'enard & 10 & 0.012066 & 0.001595 & 0.004401 \\
\hline
KS & 1  & 0.000407 & 0.000223 & 0.000339 \\
KS & 5  & 0.002140 & 0.001146 & 0.001753 \\
KS & 10 & 0.004134 & 0.002108 & 0.003505 \\
\hline
Kolmogorov & 1  & 0.018522 & 0.018065 & 0.006738 \\
Kolmogorov & 5  & 0.100077 & 0.097531 & 0.064080 \\
Kolmogorov & 10 & 0.253117 & 0.247677 & 0.165009 \\
\hline
\end{tabular}
\end{table}

\subsection{Qualitative Comparison at Native Resolution}
\begin{figure*}[!t]
\centering
\includegraphics[width=0.98\textwidth]{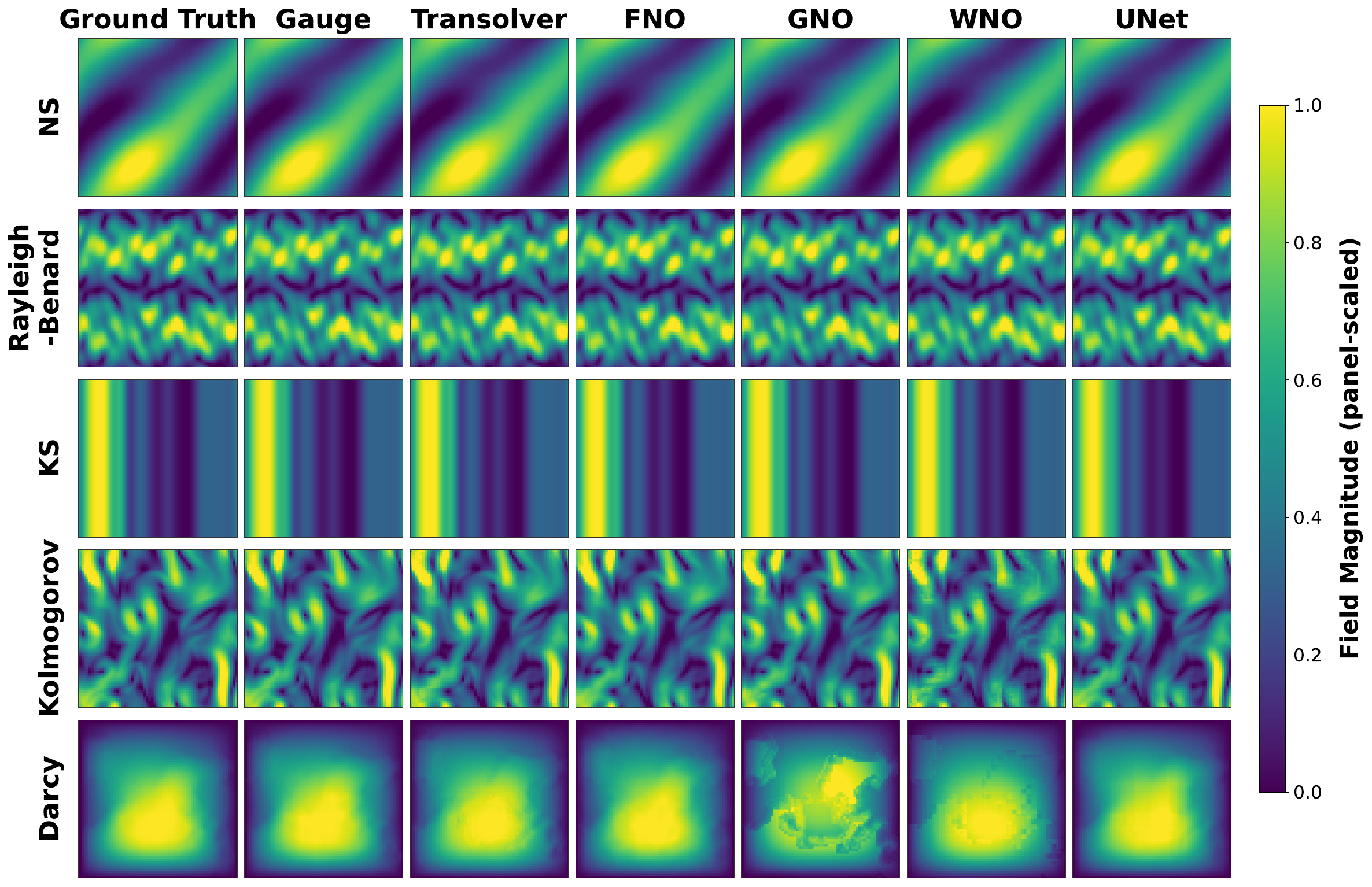}
\caption{Representative native-resolution predictions. Columns compare the ground truth, GA-AMNO, Transolver, FNO, GNO, WNO, and UNet; rows correspond to Navier-Stokes, Rayleigh-B\'enard, KS, Kolmogorov, and Darcy. Colors are independently scaled within each panel to emphasize spatial structure.}
\label{fig:qualitative_main_comparison}
\end{figure*}

\autoref{fig:qualitative_main_comparison} complements the aggregate errors in \autoref{tab:main_results_revised} with one representative prediction at the native evaluation resolution. GA-AMNO preserves the dominant spatial structures across all five PDE families, including localized flow patterns and heterogeneous Darcy responses. Each panel is scaled independently to $[0,1]$ for structural comparison, so color should not be interpreted as a shared physical amplitude across models or datasets. The one-dimensional KS profile is repeated vertically only to maintain a common field-map layout; quantitative conclusions remain those of \autoref{tab:main_results_revised}.

\section{Notation}
\label{app:notation}

The following notation summarizes the symbols used in the method,
evaluation, and theoretical analysis. Superscripts $r$ and $a$ denote
reference and adaptive discretizations, respectively. A prime denotes
a quantity after an admissible change of local representation convention.

\centerline{\bf PDE fields and operator mappings}
\bgroup
\def\arraystretch{1.5}
\begin{tabular}{p{1.25in}p{3.25in}}
$\displaystyle \Omega$
& Spatial domain. \\

$\displaystyle d$
& Spatial dimension. \\

$\displaystyle \mathcal{F}_{\mathrm{PDE}}$
& Governing differential operator. \\

$\displaystyle \boldsymbol{\mu}$
& Physical parameters, coefficient fields, or forcing terms. \\

$\displaystyle \mathcal{A}$
& Input function space of the PDE solution operator. \\

$\displaystyle \mathcal{U}$
& Output solution function space of the PDE solution operator. \\

$\displaystyle \mathbf{a}$
& Temporal input history or steady conditioning field. \\

$\displaystyle \mathcal{U}_t$
& Input history ending at time $t$. \\

$\displaystyle \mathbf{v}$
& Latest input state or conditioning field. \\

$\displaystyle \mathbf{u}$
& Ground-truth physical field. \\

$\displaystyle \widehat{\mathbf{u}}$
& Predicted physical field. \\

$\displaystyle \mathcal{G}^{\dagger}$
& Ground-truth PDE solution operator. \\

$\displaystyle \mathcal{G}_{\theta}$
& Learned neural operator parameterized by $\theta$. \\
\end{tabular}
\egroup
\vspace{0.25cm}

\centerline{\bf Adaptive discretization and state encoding}
\bgroup
\def\arraystretch{1.5}
\begin{tabular}{p{1.25in}p{3.25in}}
$\displaystyle \mathcal{X}^{r}$
& Reference discretization or canonical grid. \\

$\displaystyle \mathbf{X}^{r}$
& Canonical coordinate tensor used by the grid-based branches. \\

$\displaystyle \mathcal{X}^{a}$
& Input-dependent adaptive discretization. \\

$\displaystyle \mathbf{x}_{i}^{r}$
& Coordinate of reference node $i$. \\

$\displaystyle \mathbf{x}_{i}^{a}$
& Coordinate of adaptive node $i$. \\

$\displaystyle \mathbf{P}$
& Encoded state descriptor or physics-proxy feature map. \\

$\displaystyle \mathbf{p}_{i}$
& State descriptor at node $i$. \\

$\displaystyle Q_g$
& Gradient-magnitude physical indicator. \\

$\displaystyle Q_{\Delta}$
& Laplacian-magnitude physical indicator. \\

$\displaystyle Q_E$
& Local energy indicator. \\

$\displaystyle Q_{\omega}$
& Vorticity-related indicator. \\

$\displaystyle \widetilde{\mathbf{Q}}(\mathbf{x})$
& Concatenated normalized physical-indicator vector at location $\mathbf{x}$. \\

$\displaystyle I(\mathbf{x})$
& Predicted scalar importance map at spatial location $\mathbf{x}$. \\

$\displaystyle \mathcal{N}(\cdot)$
& Per-sample channel-normalization operator. \\

$\displaystyle \eta_{\theta}$
& Convolutional importance-map predictor. \\

$\displaystyle \sigma$
& Sigmoid activation used for importance prediction. \\
\end{tabular}
\egroup
\vspace{0.25cm}

\centerline{\bf Mesh dimensions and adaptive movement}
\bgroup
\def\arraystretch{1.5}
\begin{tabular}{p{1.25in}p{3.25in}}
$\displaystyle N$
& Number of nodes or scalar field entries, according to context. \\

$\displaystyle C$
& Number of physical channels. \\

$\displaystyle C_h$
& Hidden feature width. \\

$\displaystyle I_i$
& Learned importance value at node $i$. \\

$\displaystyle w_i$
& Positive mean-normalized importance weight. \\

$\displaystyle \mathbf{o}_i$
& Bounded raw displacement direction. \\

$\displaystyle \Delta\mathbf{x}_i$
& Importance-modulated node displacement. \\

$\displaystyle \delta_{\max}$
& Maximum allowed displacement magnitude. \\

$\displaystyle \psi_{\theta}$
& Convolutional head that predicts the bounded raw node displacement. \\

$\displaystyle I_{\min}$
& Positive lower bound used in importance modulation. \\

$\displaystyle s_I$
& Scale controlling the range of importance modulation. \\

$\displaystyle \Pi_{\Omega}$
& Projection that keeps displaced nodes inside the spatial domain $\Omega$. \\

$\displaystyle \mathbf{J}_i^a$
& Local adaptive-mesh Jacobian. \\
\end{tabular}
\egroup
\vspace{0.25cm}

\centerline{\bf Gauge-connection parameterization}
\bgroup
\def\arraystretch{1.5}
\begin{tabular}{p{1.25in}p{3.25in}}
$\displaystyle \mathbf{e}_{ij}$
& Edge descriptor for directed edge $j\rightarrow i$. \\

$\displaystyle r$
& Rank of the low-rank transport correction. \\

$\displaystyle \mathbf{d}_{ij}^{(\ell)}$
& Diagonal transport factors at layer $\ell$. \\

$\displaystyle \mathbf{U}_{ij}^{(\ell)}$
& Left low-rank transport factor. \\

$\displaystyle \mathbf{V}_{ij}^{(\ell)}$
& Right low-rank transport factor. \\

$\displaystyle \gamma_{ij}^{(\ell)}$
& Gate controlling the low-rank correction. \\

$\displaystyle \mathbf{T}_{i\leftarrow j}^{(\ell)}$
& Learned source-to-target feature transport. \\
\end{tabular}
\egroup
\vspace{0.25cm}

\centerline{\bf Attention and transported aggregation}
\bgroup
\def\arraystretch{1.5}
\begin{tabular}{p{1.25in}p{3.25in}}
$\displaystyle a_{ij}^{(\ell)}$
& Unnormalized edge-attention score. \\

$\displaystyle \alpha_{ij}^{(\ell)}$
& Softmax-normalized edge-attention weight. \\

$\displaystyle \mathbf{h}_{i}^{(\ell)}$
& Feature of node $i$ at layer $\ell$. \\

$\displaystyle \mathbf{m}_{i}^{(\ell)}$
& Aggregated transported message at node $i$. \\

$\displaystyle \mathcal{N}(i)$
& Neighborhood of node $i$. \\

$\displaystyle \rho_{\theta}^{(\ell)}$
& Learned residual feature-update network at layer $\ell$. \\

$\displaystyle L_g$
& Number of low-rank gauge transport layers. \\

$\displaystyle \epsilon$
& Positive numerical-stabilization constant. \\
\end{tabular}
\egroup
\vspace{0.25cm}

\centerline{\bf Mesh-to-grid reconstruction}
\bgroup
\def\arraystretch{1.5}
\begin{tabular}{p{1.25in}p{3.25in}}
$\displaystyle \mathbf{y}_q^r$
& Coordinate of canonical output point $q$. \\

$\displaystyle \mathcal{N}_b(q)$
& Adaptive-node neighborhood used for reconstruction. \\

$\displaystyle \kappa_{qi}$
& Interpolation weight from adaptive node $i$ to output point $q$. \\

$\displaystyle \tau_b$
& Distance temperature of the mesh-to-grid bridge. \\

$\displaystyle \mathbf{z}_q$
& Reconstructed feature at canonical point $q$. \\

$\displaystyle \mathbf{F}_{\mathrm{grid}}$
& Canonical grid feature tensor. \\
\end{tabular}
\egroup
\vspace{0.25cm}

\centerline{\bf Residual prediction and differential operators}
\bgroup
\def\arraystretch{1.5}
\begin{tabular}{p{1.25in}p{3.25in}}
$\displaystyle \boldsymbol{\delta}_{\mathrm{raw}}$
& Residual predicted by the local branch. \\

$\displaystyle \boldsymbol{\delta}_{\mathrm{spec}}$
& Residual predicted by the spectral branch. \\

$\displaystyle \boldsymbol{\delta}_1$
& Sum of local and spectral residuals. \\

$\displaystyle \mathbf{u}_{\mathrm{base}}$
& Provisional physical field used for differential correction. \\

$\displaystyle \mathbf{Z}_{\mathrm{phy}}$
& Input tensor of the differential correction head. \\

$\displaystyle \boldsymbol{\delta}$
& Final corrected residual. \\

$\displaystyle D_x$
& Discrete derivative in the first spatial direction. \\

$\displaystyle D_y$
& Discrete derivative in the second spatial direction. \\

$\displaystyle \Delta$
& Discrete Laplacian operator. \\

$\displaystyle \mathcal{F}_{\mathrm{DFT}}$
& Orthonormal discrete Fourier transform. \\
\end{tabular}
\egroup
\vspace{0.25cm}

\centerline{\bf Feature-compatibility evaluation}
\bgroup
\def\arraystretch{1.5}
\begin{tabular}{p{1.25in}p{3.25in}}
$\displaystyle \mathcal{E}_b$
& Directed-edge set in geometric-mismatch group $b$. \\

$\displaystyle E_{\mathrm{src}}^{(b)}$
& Source-to-target feature discrepancy in group $b$. \\

$\displaystyle E_{\mathrm{msg}}^{(b)}$
& Transported-message-to-target discrepancy in group $b$. \\

$\displaystyle G_{\mathrm{comp}}^{(b)}$
& Relative compatibility gain produced by transport. \\

$\displaystyle \mathbf{g}_i$
& Local discretization context at node $i$. \\

$\displaystyle \mathbf{z}_i$
& Latent physical content at node $i$. \\

$\displaystyle \mathbf{R}(\mathbf{g}_i)$
& Representation map induced by local context $\mathbf{g}_i$. \\
\end{tabular}
\egroup
\vspace{0.25cm}

\centerline{\bf Local representation analysis}
\bgroup
\def\arraystretch{1.5}
\begin{tabular}{p{1.25in}p{3.25in}}
$\displaystyle \mathfrak{G}$
& Admissible group of local representation changes. \\

$\displaystyle \mathfrak{D}$
& Family of admissible nondegenerate discretization contexts. \\

$\displaystyle \mathcal{O}(\mathbf{z}_i)$
& Admissible representation set of physical content $\mathbf{z}_i$. \\

$\displaystyle \mathbf{S}_i$
& Local representation action at node $i$. \\

$\displaystyle \overline{\mathbf{m}}_i$
& Message produced by uncorrected direct aggregation. \\

$\displaystyle \mathbf{W}$
& Shared linear map in direct aggregation. \\
\end{tabular}
\egroup
\vspace{0.25cm}

\centerline{\bf Connection consistency and ideal transport}
\bgroup
\def\arraystretch{1.5}
\begin{tabular}{p{1.25in}p{3.25in}}
$\displaystyle \delta_{\alpha,i}$
& Attention-consistency defect at node $i$. \\

$\displaystyle \varepsilon_{ij}$
& Connection-consistency defect on edge $j\rightarrow i$. \\

$\displaystyle \mathbf{T}_{i\leftarrow j}^{\star}$
& Ideal source-to-target transport matrix. \\

$\displaystyle \mathbf{T}^{\star}$
& Ideal transport matrix for a representative edge in the approximation analysis. \\

$\displaystyle \mathbf{D}$
& Diagonal component of an ideal transport. \\

$\displaystyle \mathbf{E}$
& Non-diagonal remainder $\mathbf{T}^{\star}-\mathbf{D}$. \\

$\displaystyle \sigma_k(\mathbf{E})$
& The $k$th singular value of $\mathbf{E}$. \\
\end{tabular}
\egroup
\vspace{0.25cm}

\centerline{\bf Mesh and operator stability constants}
\bgroup
\def\arraystretch{1.5}
\begin{tabular}{p{1.25in}p{3.25in}}
$\displaystyle \delta_{\mathcal{X}}$
& Maximum coordinate change between two meshes. \\

$\displaystyle L_m$
& Local Lipschitz constant for message variation. \\

$\displaystyle \Phi_{\ell}$
& Complete feature-update map at layer $\ell$. \\

$\displaystyle b_{\ell}$
& Representation-consistency defect at layer $\ell$. \\

$\displaystyle K_{\ell}$
& Lipschitz constant of layer $\ell$. \\

$\displaystyle \mathcal{D}$
& Canonical reconstruction and prediction map. \\

$\displaystyle K_{\mathrm{out}}$
& Lipschitz constant of the output map. \\

$\displaystyle \mathbf{S}_{\mathrm{out}}$
& Prescribed representation action on the output. \\

$\displaystyle b_{\mathrm{out}}$
& Representation-consistency defect of the output map. \\
\end{tabular}
\egroup

\section*{AI Use Statement}

In this work, we used generative AI tools to assist with language polishing and limited programming support. All scientific ideas, methodology, experimental design, implementation decisions, analysis, and conclusions were developed and verified by the authors. All AI-assisted content was carefully reviewed and revised by the authors. We take full responsibility for the final content of this work.

\end{document}